\documentclass[10pt,journal,cspaper,compsoc]{IEEEtran}
\usepackage{algorithmic}
\usepackage{algorithm}
\usepackage{amsmath,epsfig}
\usepackage{amsmath,bm}
\usepackage{array}
\usepackage{graphics}
\usepackage{graphicx}
\usepackage{amssymb}
\usepackage{amsthm}
\usepackage{float}
\usepackage{subfigure}
\usepackage{booktabs}
\usepackage{ccaption}
\usepackage{extarrows}
\usepackage{mathtools}
\usepackage{caption}
\ifCLASSOPTIONcompsoc
\else
\fi

\ifCLASSINFOpdf
\else
\fi

\begin{document}
%
\title{Online Bagging and Boosting for Imbalanced Data Streams}
%
%
%
%

\author{Boyu~Wang
        and~Joelle~Pineau
\IEEEcompsocitemizethanks{\IEEEcompsocthanksitem The authors are with the School of Computer Science,
McGill University, McConnell Engineering Bldg, 3480 University Street, Montreal, QC H3A 0E9, Canada.\protect\\
E-mail: boyu.wang@mail.mcgill.ca; jpineau@cs.mcgill.ca.}
\thanks{}}

\IEEEcompsoctitleabstractindextext{%
\begin{abstract}
While both cost-sensitive learning and online learning have been studied extensively, the effort in simultaneously dealing with these two issues is limited. Aiming at this challenge task, a novel learning framework is proposed in this paper. The key idea is based on the fusion of online ensemble algorithms and the state of the art batch mode cost-sensitive bagging/boosting algorithms. Within this framework, two separately developed research areas are bridged together, and a batch of theoretically sound online cost-sensitive bagging and online cost-sensitive boosting algorithms are first proposed. Unlike other online cost-sensitive learning algorithms lacking theoretical analysis of asymptotic properties, the convergence of the proposed algorithms is guaranteed under certain conditions, and the experimental evidence with benchmark data sets also validates the effectiveness and efficiency of the proposed methods.

\end{abstract}

}

\maketitle

\IEEEdisplaynotcompsoctitleabstractindextext
\IEEEpeerreviewmaketitle

\section{Introduction}

\IEEEPARstart{I}{n} many real world applications, such as medical diagnosis \cite{park2011seizure}, network intrusion detection \cite{cieslak2006combating}, spam filtering \cite{fawcett2003vivo}, the distribution between the classes of examples is not well balanced. In a binary classification scenario, such as automatically detecting whether an incoming individual is an intruder or not, we may have many more negative examples (i.e., non-intruders) than positive examples (i.e., intruders). The class with the most examples is typically referred to as the majority class, and the other one as the minority class. Conventional learning algorithms in machine learning and data mining typically do not work well for imbalanced class problems since their objective is to minimize the overall error rate, which implicitly treats all misclassification costs equally. As a result, these algorithms may produce trivial results, typically classifying all test examples as negative. Additionally, it is often the case that the positive (minority) class is the one of greater interest. For example, the expected cost of missing an intruder may be much higher than the cost of mis-identifying a non-intruder. For this reason, the class imbalance problem \cite{he2009learning} has often been formulized within the cost-sensitive learning framework \cite{elkan2001foundations}.

The class imbalance problem gets more challenging in the context of learning from data streams, which unfortunately is required in many practical classification problems. For example, in epileptic EEG/ECoG data analysis \cite{gotman1990automatic}, one challenge is that seizures are relatively rare, compared to non-seizure brain activities, and the cost of missing a seizure is much higher than detecting seizure-free signal into a seizure. Meanwhile, in the clinical scenario, the EEG/ECoG data is usually collected incrementally over time, and the seizure detection system must respond to the signals in real time. Even though the entire data can be received from the start and the classifier can be trained in batch mode, it is still favorable if the model can adapt online to new data. Such data may come from the same subject but from different time, or even from different subjects. This is because the size of the entire data set can be too large to fit in memory or to train a classifier at once. Moreover, since the positive examples (seizures) are rather rare, each of them should be made full use of. However, in the batch learning scenario, the trained classifier is static and is not updated in the testing phase. Therefore, this problem should also be formulated within the incremental learning framework.\footnote{In this paper, incremental learning is referred to the general technique learning from data streams, and online learning is characterized responding immediately to a new instance and then discarding it, except for SMOTE-based online learning algorithms, the positive examples are stored to create synthetic samples.} The main challenge of such a problem is how to perform well even with very few positive examples at the early stage of model building, and perform better as more examples are available, motivating us to develop effective incremental learning algorithms dealing with class imbalance problem.

Many techniques have been proposed to deal with class imbalance problem, among which ensemble learning approaches\footnote{In this paper, we restrict ourselves to bagging and boosting approaches.} have been proved more effective than other algorithms \cite{galar2012review}. On the other hand, many methods for learning from data streams are also available in the literature on incremental/online learning. Although these two issues have been well studied independently in the past decades, the research on simultaneously solving both of the problems is limited. In this paper, an online learning with class imbalance framework is proposed. Our main contributions can be summarized as follows.

\begin{enumerate}
\item [1.]We propose an online cost-sensitive ensemble learning framework, which generalizes a batch of widely used bagging and boosting based cost-sensitive learning algorithms to their online version.
\item [2.]We theoretically analyze the consistency between the proposed algorithms and their batch mode counterparts, showing that under certain conditions, as the number of examples approaches infinity, the models generated by online cost-sensitive ensemble learning algorithms converge to that of batch cost-sensitive ensembles.
\item [3.] Two long separate research areas, cost-sensitive learning and incremental learning, are bridged together in this paper. As meta-learning techniques, the proposed framework can convert any existing cost-insensitive online learning algorithm into cost-sensitive one.
\item [4.] With some straightforward modifications, the proposed algorithms can also deal with the concept drift problem in non-stationary environments.
\end{enumerate}

The performance of the proposed online cost-sensitive bagging and boosting algorithms is evaluated on the UCI data sets, and a comprehensive comparison study of online and batch cost-sensitive ensemble learning algorithms is presented. It should be emphasized that we mainly focus on the consistency between the proposed methods and their batch mode counterparts. The comparison among different ensembles for class imbalance or deciding the best one, which can be found in \cite{galar2012review}, \cite{khoshgoftaar2011comparing}, is beyond the scope of this paper. The remainder of this paper is organized as follows. Section 2 briefly reviews the related work of class imbalance learning and online learning. Section 3 presents technical background of our work. The online cost-sensitive ensemble learning framework is proposed in Section 4 followed by the theoretical analysis in Section 5. The experimental results are reported in Section 6, and some discussions are provided in Section 7. Section 8 extends the proposed framework to non-stationary environments. Section 9 completes the paper with conclusions.

\section{Related Work}

Existing approaches to dealing with class imbalanced problem can be roughly categorized into algorithm level approaches and data level approaches. The former directly modifies the traditional algorithms to achieve cost sensitivity by taking different misclassification costs into consideration when designing the algorithms such that the misclassification cost of positive examples is higher than that of negative ones. The goal of a classifier is to minimize the cost instead of classification error, and therefore the classification algorithms will bias towards the small class. For example, cost-sensitive SVM can be derived by kernel modification \cite{wu2003adaptive}, biased penalty \cite{lin2002support}, or loss function modification \cite{masnadi2010risk}. For decision tree, the cost sensitivity can be introduced by probabilistic estimate calibration \cite{zadrozny2001learning} or using different pruning methods \cite{drummond2000exploiting}. At the data level, a data preprocessing step is added to rebalance the class distribution by undersampling the negative class, oversampling the positive class, or creating synthetic positive examples \cite{chawla2002smote}, \cite{chawla2004editorial}, and then conventional cost-insensitive algorithms are applied on the rebalanced data sets. Such approaches can also be regarded as meta-learning techniques, that is, they can convert cost-insensitive algorithms into cost-sensitive ones without modifying them. One advantage of this category of approaches is that it is independent of specific classifiers and therefore it is applicable to any existing cost-insensitive learning algorithm. The sampling techniques can be further incorporated with ensemble learning algorithms \cite{polikar2006ensemble} and have received much attention recently due to their better performance on imbalanced data sets \cite{galar2012review}. Ensemble based techniques for class imbalance problem inherit the good properties of ensemble learning algorithms, improving the generalization ability of learning algorithms via bias or/and variance reduction, as well as achieving cost sensitivity by sampling techniques. In addition, traditional ensemble learning algorithms themselves have sampling step in each iteration. Therefore, little extra learning cost is added when embedding the resampling step to rebalance the data set. 

As for learning from data streams, some algorithms are naturally or can be easily extended to incremental version, including \textit{k}-NN, naive Bayes classifier, binary linear discriminant analysis (LDA), quadratic discriminant analysis (QDA), and so on. In addition, the incremental/online versions of more sophisticated algorithms have been proposed in the literature, including but not limited to decision trees \cite{utgoff1997decision}, random forests \cite{saffari2009line}, \cite{denil2013consistency}
, multi-class LDA \cite{liu2009least}, \cite{kim2007incremental}, logistic regression \cite{langford2009sparse}, support vector machines \cite{laskov2006incremental}, \cite{cauwenberghs2001incremental}, and other kernel methods \cite{kivinen2004online}, \cite{hoi2013online}. Besides the base learning algorithms, the online versions of ensemble learning techniques, bagging and boosting, were also derived in \cite{oza2001online} by approximating binominal distribution using a Poisson distribution.

\section{Technical Background}
In this section, we briefly review the standard bagging and boosting algorithms, as well as their online versions and cost-sensitive versions, which motivate the proposed online cost-sensitive ensemble framework.
\subsection{Standard bagging and boosting}

Ensemble learning algorithms work by combining the outputs of multiple base learners. The basic idea here is to improve the generalization ability of individual classifiers by training them on different data sets, which is motivated by bias-variance trade-off \cite{bishop2006pattern}. Averaging the outputs of base models tends to cancel the variance component or/and reduce the bias. On the other hand, to achieve good performance, the diversity, which is usually introduced by presenting different training data to different base models among classifiers, is an important characteristic \cite{kuncheva2003measures}. Different approaches to averaging base models and obtaining desired diversity distinguish different ensemble methods. Two representative techniques among them are Bagging \cite{breiman1996bagging} and Boosting (AdaBoost) \cite{freund1995desicion}.

Given a data set \(S=\{(x_1,y_1),\dots,(x_N,y_N)\}\) of size $N$, where \(x_n\in X\), \(y_n\in Y = \{0,1\}\), \(M\) base models \(h_m\), bagging constructs \(M\) classifiers with bootstrap replicas \(\{S_m\}\) of \(S\), where \(S_m\) is obtained by drawing examples from original the date set \(S\) with replacement, usually having the same number of examples as \(S\). The diversity among the classifiers is introduced by independently constructing different subsets of the original data set. After constructing ensembles, the prediction of the class of a new example is given by majority voting.  The pseudo-code of Bagging is shown in Alg.~\ref{alg:bagging}.

\begin{algorithm}[h]\footnotesize
\caption{Bagging Algorithm}
\label{alg:bagging}

\textbf{Input:} \(S,M\)
  \begin{algorithmic}[1]
    \FOR{\(m=1,\dots,M\)}
        \STATE $S_m=Sample\_with\_replacement(S,N)$
        \STATE Train a base learner $h_m\to Y$ using $S_m$
    \ENDFOR
  \end{algorithmic}
  \raggedright\textbf{Output:} \(H(x)=\mathop{\arg \max}\limits_{y\in Y}\sum\limits _{m=1}^M I(h_m(x)=y)\)

\end{algorithm}

AdaBoost is another widely used ensemble learning algorithm. Unlike Bagging, which treats all of examples equally, AdaBoost focuses more on difficult examples. In particular, Adaboost sequentially constructs a series of base learner in such a way that examples that are misclassified by current base learner $h_m$ are given more weights in the training set for the following learner $h_{m+1}$, whereas the correctly classified examples are given less weights. More specifically, the weights of all examples are initially equal, and then examples misclassified by $h_m$ are given half the total weight for the following learner $h_{m+1}$, and the correctly classified examples are given the remaining half of the total weights. The pseudo-code of AdaBoost is shown in Alg.~\ref{alg:adaboost}.

It should be emphasized that by using the update rule in step 6, the normalization step is avoided. That is, the summation of $D_{m+1}$ will remain one after each update without normalization, which is crucial to designing the online boosting algorithm \cite{oza2001online} and the proposed online cost-sensitive boosting algorithms, since the normalization factor is unavailable during the online learning process.  After update, the reweighted examples can be either directly used to train the next base learner, or first resampled according to the weights and then the unweighted samples are used to train the base learner. In this paper, all boosting techniques are implemented by resampling. The reasons are three folds: First, sampling based boosting algorithms are consistent with bagging techniques. Second, they are also consistent with their online counterparts introduced later, which simulate sampling with replacement by using Poisson distribution. Finally, reweighting on training set is not applicable to all learning algorithms.

\begin{algorithm}[h]\footnotesize
\caption{Boosting (AdaBoost) Algorithm}
\label{alg:adaboost}
  \textbf{Input:} \(S,M\)
  \begin{algorithmic}[1]
  \STATE Initialize $D_1(n)=\frac {1}{N}$ for all $n\in \{1,\dots,N\}$
    \FOR{\(m=1,\dots,M\)}
        \STATE Train a base learner $h_m\to Y$ using $S$ with distribution $D_m$
        \STATE $\epsilon _m = \sum\limits _{n=1}^N D_m(n)I(h_m(x_n)\neq y_n)$
        \FOR{\(n=1,\dots,N\)}
            \STATE \(D_{m+1}(n)=D_m(n)\times \begin{cases}
            \frac {1}{2(1-\epsilon _m)}, \hfill\quad h_m(x_n)=y_n \\
            \frac {1}{2\epsilon _m},\hfill\quad h_m(x_n)\neq y_n
            \end{cases} \)
        \ENDFOR
    \ENDFOR
  \end{algorithmic}
  \raggedright\textbf{Output:} \(H(x)=\mathop{\arg \max}\limits_{y\in Y}\sum\limits _{m=1}^M log(\frac{1-\epsilon_m}{\epsilon_m})I(h_m(x)=y)\)
\end{algorithm}

\subsection{Online bagging and boosting}

The framework of online ensemble learning algorithms \cite{oza2001online}, \cite{oza2001onlinephd} is inspired by the observation that the binomial distribution $Binomial(p,N)$ can be approximated by a Poisson distribution $Poisson(\lambda)$ with $\lambda =Np$ as $N\to \infty$, where the probability of success $p$ in the binomial distribution is equivalent to $D(n)$ in bagging and boosting algorithms. For example, since $D(n)=\frac{1}{N}$ for all examples in bagging algorithm, the uniform sampling with replacement of bagging algorithms can be approximated by $Poisson(1)$. For online boosting, $\lambda$ can be computed by tracking the total weights of correctly classified and misclassified examples for each base learner ($\lambda _m^{SC}, \lambda _m^{SW}$).The online bagging and boosting algorithms are described in Alg.~\ref{alg:onlinebagging} and Alg.~\ref{alg:onlineadaboost} respectively. For more detailed description of the online ensemble learning framework, we refer the reader to \cite{oza2001online}, \cite{oza2001onlinephd}.

\begin{algorithm}[h]\footnotesize
\caption{Online Bagging Algorithm}
\label{alg:onlinebagging}
  \textbf{Input:} \(S,M\)
  \begin{algorithmic}[1]
    \FOR{\(n=1,\dots,N\)}
        \FOR{\(m=1,\dots,M\)}
            \STATE Let $k\sim Poisson(1)$
            \STATE Do $k$ times \\
            \quad Train the base learner $h_m\to Y$ using $(x_n,y_n)$
        \ENDFOR
    \ENDFOR
  \end{algorithmic}
  \raggedright\textbf{Output:} \(H(x)=\mathop{\arg \max}\limits_{y\in Y}\sum\limits _{m=1}^M I(h_m(x)=y)\)
\end{algorithm}

\begin{algorithm}[h]\footnotesize
\caption{Online Boosting (AdaBoost) Algorithm}
\label{alg:onlineadaboost}
  \textbf{Input:} \(S,M\)
  \begin{algorithmic}[1]
  \STATE Initialize $\lambda_m^{SC} = 0, \lambda_m^{SW} = 0$ for all $m\in \{1,\dots,M\}$
    \FOR{\(n=1,\dots,N\)}
        \STATE Set $\lambda = 1$
        \FOR{\(m=1,\dots,M\)}
            \STATE Let $k\sim Poisson(\lambda)$
            \STATE Do $k$ times
            \STATE \quad Train the base learner $h_m\to Y$ using $(x_n,y_n)$
            \IF{$h_m(x_n)=y_n$}
                \STATE $\lambda_m^{SC} \leftarrow \lambda_m^{SC} +\lambda,\epsilon_m\leftarrow \frac{\lambda_m^{SW}}{\lambda_m^{SC} + \lambda_m^{SW}}, \lambda \leftarrow \frac{\lambda}{2(1-\epsilon_m)}$
            \ELSE
                \STATE $\lambda_m^{SW} \leftarrow \lambda_m^{SW} +\lambda,\epsilon_m\leftarrow \frac{\lambda_m^{SW}}{\lambda_m^{SC} + \lambda_m^{SW}}, \lambda \leftarrow \frac{\lambda}{2\epsilon_m}$
            \ENDIF
        \ENDFOR
    \ENDFOR
  \end{algorithmic}
  \raggedright\textbf{Output:} \(H(x)=\mathop{\arg \max}\limits_{y\in Y}\sum\limits _{m=1}^M log(\frac{1-\epsilon_m}{\epsilon_m})I(h_m(x)=y)\)
\end{algorithm}

\subsection{Cost-sensitive bagging and boosting}

Cost-sensitive ensemble learning is a meta-technology that takes different misclassification costs into consideration via biased resampling/reweighting methods before each iteration of bagging/boosting. As a result, cost-sensitive ensemble learning algorithms can turn any cost-insensitive classifier into a cost-sensitive one with little extra learning cost, while preserving the good properties of standard ensemble learning algorithms (i.e., bias and/or variance reduction). The main difference between these techniques lies in the choice of resampling mechanism. From this point, we briefly review six popular cost-sensitive ensemble learning algorithms: UnderOverBagging \cite{wang2009diversity}, SMOTEBagging \cite{wang2009diversity}, AdaC2 \cite{sun2007cost}, CSB2 \cite{ting2000comparative}, RUSBoost \cite{seiffert2010rusboost}, and SOMOTEBoost \cite{chawla2003smoteboost}. The derivation of their online extensions is the main contribution of this paper and is described in the next section.

Given an imbalanced data set of $N^+$ minority class $S^+$ and $N^-$ majority class $S^-$, one straightforward approach to implement bagging-based ensemble learning algorithms is to undersample the majority class or oversample the minority class, which gives UnderBagging and OverBagging respectively. UnderOverBagging \cite{wang2009diversity} is a uniform approach combining both UnderBagging and OverBagging. In addition, the resampling rate ($a\%$) can be also varied over the bagging iterations, which further boosts the diversity among the base learners.

Alg.~\ref{alg:underoverbagging} shows the pseudo-code of UnderOverBagging, from which it can be observed that the sampling method is gradually switched from undersampling the majority class to oversampling the minority class. The number of training examples for the first base learner is lower than the last one. Since both UnderBagging and OverBagging can be regarded as special cases of UnderOverBagging, only UnderOverBagging is investigated in this paper.

\begin{algorithm}[h]\footnotesize
\caption{Batch UnderOverBagging Algorithm}
\label{alg:underoverbagging}
  \textbf{Input:} \(S,M,C>1\)
  \begin{algorithmic}[1]
    \FOR{\(m=1,\dots,M\)}
        \STATE $a=\frac{100m}{M}$
        \STATE $S_{train}^-=Sample\_with\_replacement(S^-,N^-a\%)$
        \STATE $S_{train}^+=Sample\_with\_replacement(S^+,CN^+a\%)$
        \STATE $S_m =S_{train}^+ +S_{train}^-$
        \STATE Train a base learner $h_m\to Y$ using $S_m$
    \ENDFOR
  \end{algorithmic}
  \raggedright\textbf{Output:} \(H(x)=\mathop{\arg \max}\limits_{y\in Y}\sum\limits _{m=1}^M I(h_m(x)=y)\)
\end{algorithm}

In SMOTEBagging \cite{wang2009diversity} (Alg.~\ref{alg:smotebagging}), the negative class is sampled with replacement at rate $100\%$ (i.e., $N^-$ negative examples are generated), while $CN^+$ positive examples are generated for each base learner, among which $a\%$ of them are created by resampling and the rest of the examples are created by the synthetic minority over-sampling technique (SMOTE) \cite{chawla2002smote} shown in Alg.~\ref{alg:smote}. The main idea of SMOTE is to generating more new synthetic examples by interpolating the positive examples. As a result, all of the base learners are trained on a more balanced and diverse data set. The diversity is further boosted by varying $a\%$ so that the ratios of bootstrap replicates and synthetic examples generated by SMOTE varies over the bagging iterations.

\begin{algorithm}[h]\footnotesize
\caption{Batch SMOTEBagging Algorithm}
\label{alg:smotebagging}
  \textbf{Input:} \(S,M,k,C>1\)
  \begin{algorithmic}[1]
    \FOR{\(m=1,\dots,M\)}
        \STATE $a=\frac{100m}{M}$
        \STATE $S_{train}^-=Sample\_with\_replacement(S^-,N^-)$
        \STATE $S_{train}^+=Sample\_with\_replacement(S^+,CN^+a\%)$
        \STATE $S_{S}^+=SMOTE(S^+,C(1-a\%),k)$
        \STATE $S_m =S_{train}^+ +S_{train}^- + S_{S}^+$
        \STATE Train a base learner $h_m\to Y$ using $S_m$
    \ENDFOR
  \end{algorithmic}
  \raggedright\textbf{Output:} \(H(x)=\mathop{\arg \max}\limits_{y\in Y}\sum\limits _{m=1}^M I(h_m(x)=y)\)

\end{algorithm}

\begin{algorithm}[h]\footnotesize
\caption{Batch SMOTE Algorithm}
\label{alg:smote}
  \textbf{Input:} \(S,T,k\)
  \begin{algorithmic}[1]
    \FOR{\(n=1,\dots,N\)}
        \FOR{\(i=1,\dots,T\)}
            \STATE Randomly choose one of the $k$ nearest neighbors of $x_n$, $x_n'$
            \STATE Calculate the difference between $x_n$ and $x_n'$
            \STATE Generate a random number $\gamma$ between 0 and 1
            \STATE Create a synthetic instance: \\ $x_{S,n}^i = x_n + \gamma (x_n'-x_n)$
        \ENDFOR
    \ENDFOR
  \end{algorithmic}
  \raggedright\textbf{Output:} $TN$ synthetic instances $\{x_{S,1}^1,\dots,x_{S,N}^T\}$
\end{algorithm}

In \cite{wang2009diversity}, the sampling rate $C$ for both UnderOverBagging and SMOTEBagging was set as $C=\frac{N^-}{N^+}$ to obtain balanced class distribution. Howerver, it has been reported that the optimal class ratio need not be $1:1$ [41]. To eliminate this effect, we vary the resampling rate to generate an ROC to compare different cost-sensitive algorithms and their online counterparts.

AdaC2 \cite{sun2007cost} is a boosting algorithm that takes the different misclassification costs into consideration when calculating the classifier weights, and updates the sample weights. In particular, AdaC2 increases more weights on the misclassified positive class, compared to misclassified negative ones. Similarly, it decreases less the weights on  correctly classified positive examples compared to correctly classified negative examples. The pseudo-code for AdaC2 is shown in Alg.~\ref{alg:AdaC2}, where $C_P$ is the cost of misclassifying a positive example as a negative one, and $C_N$ is the cost of the contrary case. In cost-sensitive learning problems, as $C_P>C_N$, it can be observed that by embedding different costs for each class into the update formula, AdaC2 puts more weight on positive examples than negative ones.

\begin{algorithm}[h]\footnotesize
\caption{Batch AdaC2 Algorithm}
\label{alg:AdaC2}
  \textbf{Input:} \(S,M,C_N,C_P\)
  \begin{algorithmic}[1]
  \STATE Initialize $D_1(n)=\frac {1}{N}$ for all $n\in \{1,\dots,N\}$
    \FOR{\(m=1,\dots,M\)}
        \STATE Train a base learner $h_m\to Y$ using $S$ with distribution $D_m$
        \STATE Calculate $\alpha_m=\frac{1}{2}log\left(\frac{\sum_{n=1}^NC_nD_m(n)I(h_m(x_n)=y_n)}{\sum_{n=1}^NC_nD_m(n)I(h_m(x_n)\neq y_n)}\right)$,
        \\ where $C_n = C_P$ if $y=1$, $C_n = C_N$ otherwise
        \FOR{\(n=1,\dots,N\)}
            \STATE $D_{m+1}(n)=\frac{C_nD_m(n)exp(-\alpha_m(2I(h_m(x_n)=y_n)-1))}{Z_m}$ \\
            where $Z_m$ is a normalization factor
        \ENDFOR
    \ENDFOR
  \end{algorithmic}
  \raggedright\textbf{Output:} \(H(x)=\mathop{\arg \max}\limits_{y\in Y}\sum\limits _{m=1}^M \alpha I(h_m(x)=y)\)
\end{algorithm}

The CSB2 algorithm \cite{ting2000comparative} (Alg.~\ref{alg:CSB2}) is a kind of compromise between AdaBoost and AdaC2. For correctly classified examples, CSB2 treats them in the same way as AdaBoost, while for misclassified examples, it does the same as AdaC2. In addition, the voting weight of each base learner in CSB2 is the same as AdaBoost. AdaC2 and CSB2 take different misclassification costs by directly modifying the weight update equations. The key feature of this category of algorithms is that the cost sensitivity is introduced by unequally treating the examples from different classes. Therefore, such approaches are categorized as cost-sensitive boosting \cite{galar2012review}.

\begin{algorithm}[h]\footnotesize
\caption{Batch CSB2 Algorithm}
\label{alg:CSB2}
  \textbf{Input:} \(S,M,C_N,C_P\)
  \begin{algorithmic}[1]
  \STATE Initialize $D_1(n)=\frac {1}{N}$ for all $n\in \{1,\dots,N\}$
    \FOR{\(m=1,\dots,M\)}
        \STATE Train a base learner $h_m\to Y$ using $S$ with distribution $D_m$
        \STATE Calculate $\alpha_m=\frac{1}{2}log\left(\frac{\sum_{n,h_m(x_n)\neq y_n}D_m(n)}{\sum_{n,h_m(x_n) = y_n}D_m(n)}\right)$,
        \FOR{\(n=1,\dots,N\)}
            \STATE $D_{m+1}(n)=\frac{C_\delta D_m(n)exp(-\alpha_m(2I(h_m(x_n)=y_n)-1))}{Z_m}$ \\
            where $Z_m$ is a normalization factor,\\
            $C_\delta = 1$ if $h_m(x_n) = y_n$, $C_\delta = C_n$ otherwise
        \ENDFOR
    \ENDFOR
  \end{algorithmic}
  \raggedright\textbf{Output:} \(H(x)=\mathop{\arg \max}\limits_{y\in Y}\sum\limits _{m=1}^M \alpha I(h_m(x)=y)\)
\end{algorithm}

Cost sensitivity can also be introduced by sampling techniques as a pre-processing step before each iteration of the standard AdaBoost algorithm, as in UnderOverBagging and SMOTEBagging, to bias boosting algorithms towards the positive class, resulting in RUSBoost \cite{seiffert2010rusboost} and SMOTEBoost \cite{chawla2003smoteboost}\footnote{AdC2, CSB2 and RUSBoost, SMOTEBoost were respectively categorized into different families in \cite{galar2012review}. Here we do not distinguish them since all of these methods can be formulated within the framework of combination of resampling techniques and conventional boosting algorithms, and we refer to all of them as cost-sensitive boosting algorithms}. These two algorithms differ in the way they undersample the negative class or oversample the positive class by SMOTE. In particular, both algorithms rebalance the class distribution before representing the training data to base learners, yet the error estimate of base learners is still measured on the original data set. The RUSBoost and SMOTEBoost algorithms are shown in Alg.~\ref{alg:rusboost} and Alg.~\ref{alg:smoteboost} respectively\footnote{RUSBoost and SMOTEBoost are originally based on AdaBoost.M2, which is a variant of AdaBoost for multi-classification problems. As for binary classification problems, AdaBoost.M2 with base learner outputing hard labels is equivalent to AdaBoost. Therefore, in order to develop online RUSBoost and SMOTEBoost, both of them are based on AdaBoost in this paper}. For each algorithm, there are at least three different implementations. Take RUSBoost for example, in RUSBoost1, the class ratio is fixed, which means the ratio of the weighted positive and negative examples are fixed, and then the modified training set are generated according to their weights. The proportion of examples of the two classes for each base learner is not necessarily fixed. The difference between RUSBoost2 and RUSBoost1 is that it is the ratio of unweighted examples of two classes that is fixed, which means that the proportion of examples of the two classes that pass through each base learner is fixed. RUSBoost3 is easier to understand: the sampling rate of negative class is $\frac{1}{C}$ of that of positive class. The three implementations of SMOTEBoost are analogous to the ones of RUSBoost. The only difference is that instead of undersampling the majority class, SMOTEBoost algorithms oversample the minority class by generating synthetic positive examples using SMOTE.

\begin{algorithm}[h]\footnotesize
\caption{Batch RUSBoost Algorithm}
\label{alg:rusboost}
  \textbf{Input:} \(S,M,C\)
  \begin{algorithmic}[1]
  \STATE Initialize $D_1(n)=\frac {1}{N}$ for all $n\in \{1,\dots,N\}$
    \FOR{\(m=1,\dots,M\)}
        \STATE Modify the distribution $D_m$ by undersampling the majority class and generate a new training set $S'$
        \STATE Train a base learner $h_m\to Y$ using $S'$
        \STATE $\epsilon _m = \sum\limits _{n=1}^N D_m(n)I(h_m(x_n)\neq y_n)$
        \FOR{\(n=1,\dots,N\)}
            \STATE \(D_{m+1}(n)=D_m(n)\times \begin{cases}
            \frac {1}{2(1-\epsilon _m)}, \hfill\quad h_m(x_n)=y_n \\
            \frac {1}{2\epsilon _m},\hfill\quad h_m(x_n)\neq y_n
            \end{cases} \)
        \ENDFOR
    \ENDFOR
  \end{algorithmic}
  \raggedright\textbf{Output:} \(H(x)=\mathop{\arg \max}\limits_{y\in Y}\sum\limits _{m=1}^M log(\frac{1-\epsilon_m}{\epsilon_m})I(h_m(x)=y)\)
  \rule{\linewidth}{.1pt}
  \textbf{RUSBoost1} (fix the class ratio)\\
  Generate a new data set $S'$ by undersampling the majority class:\\
  Randomly remove majority class examples until $CN^+$ of them left, and then renormalize the weight distribution of the remaining training examples with respect to their total sum of weights.
  \rule{\linewidth}{.1pt}
  \textbf{RUSBoost2} (fix the example distribution)\\
  Modify the distribution $D_m$ by undersampling the majority class:\\
  Generate $CN^+$ and $N^+$ majority and minority examples respectively by sampling $D_m$.
  \rule{\linewidth}{.1pt}
  \textbf{RUSBoost3} (fix the sampling rate)\\
  Modify the distribution $D_m$ by undersampling the majority class:\\
  Generate $N$ exmples according to distribution $D_m$, and then undersample majority class until $\frac{1}{C}$ of them left.

\end{algorithm}

\begin{algorithm}[h]\footnotesize
\caption{Batch SMOTEBoost Algorithm}
\label{alg:smoteboost}
  \textbf{Input:} \(S,M,C\)
  \begin{algorithmic}[1]
  \STATE Initialize $D_1(n)=\frac {1}{N}$ for all $n\in \{1,\dots,N\}$
    \FOR{\(m=1,\dots,M\)}
        \STATE Modify the distribution $D_m$ by creating synthetic examples from minority class using SMOTE
        \STATE Train a base learner $h_m\to Y$ using $S$ with distribution $D_m$
        \STATE $\epsilon _m = \sum\limits _{n=1}^N D_m(n)I(h_m(x_n)\neq y_n)$
        \FOR{\(n=1,\dots,N\)}
            \STATE \(D_{m+1}(n)=D_m(n)\times \begin{cases}
            \frac {1}{2(1-\epsilon _m)}, \hfill\quad h_m(x_n)=y_n \\
            \frac {1}{2\epsilon _m},\hfill\quad h_m(x_n)\neq y_n
            \end{cases} \)
        \ENDFOR
    \ENDFOR
  \end{algorithmic}
  \raggedright\textbf{Output:} \(H(x)=\mathop{\arg \max}\limits_{y\in Y}\sum\limits _{m=1}^M log(\frac{1-\epsilon_m}{\epsilon_m})I(h_m(x)=y)\)
  \rule{\linewidth}{.1pt}
  \textbf{SMOTEBoost1} (fix the class ratio)\\
  Modify the distribution $D_m$ by creating synthetic examples from minority class using SMOTE:\\
  Randomly generate $\frac{N^-}{C}-N^+$ synthetic examples from minority class, and then renormalize the weight distribution with respect to their total sum of weights.
  \rule{\linewidth}{.1pt}
  \textbf{SMOTEBoost2} (fix the example distribution)\\
  Modify the distribution $D_m$ by creating synthetic examples from minority class using SMOTE:\\
  Generate $N^-$ and $N^+$ majority and minority examples respectively by sampling $D_m$, and then create  $\frac{N^-}{C}-N^+$ synthetic minority examples
  \rule{\linewidth}{.1pt}
  \textbf{SMOTEBoost3} (fix the sampling rate)\\
  Modify the distribution $D_m$ by creating synthetic examples from minority class using SMOTE:\\
  Generate $N$ examples according to distribution $D_m$, create $(C-1)N'^+$ synthetic minority examples, where $N'^+$ is the number of minority examples on the $m$th iteration.

\end{algorithm}

\section{Methods}

Our proposed framework for online cost-sensitive ensemble learning is based on the combination of online ensemble methods \cite{oza2001online} and batch cost-sensitive ensembles described above. More specifically, the cost sensitivity of batch approaches is introduced by different resampling mechanism, while in the online ensembles it is achieved by manipulating the parameters of the Poisson distribution for different classes. In this section, the online extensions of all of the popular methods described in previous section are derived. Note that the proposed framework can also be applied to other batch cost-sensitive ensembles, such as RBBagging \cite{hido2009roughly}, CSRoulette \cite{sheng2007roulette}, RareBoost \cite{joshi2001evaluating}, or DataBoost-IM \cite{guo2004learning}.

The main challenge of generalizing the cost-sensitive ensemble algorithms to online mode is how to impose the cost setting on standard online ensemble algorithms. This is straightforward for some cases. In some other cases, however, it is not straightforward to embed costs into online ensembles, especially for boosting based algorithms. In particular, the weight update step in standard AdaBoost only involves the classification of each base learner without normalization, so the online boosting algorithm can be implemented by simply tracking the error and using it to update the Poisson parameter $\lambda$ for each base learner. Therefore, the key point for us is to reformulate the batch cost-sensitive boosting algorithms in a way that there is no normalization step at each iteration, and then incrementally estimate the quantities embedded with the cost setting in the online learning scenario. Also, even though the asymptotic properties of standard online ensemble algorithms have been thoroughly studied in \cite{oza2001onlinephd}, the theoretical properties in online cost-sensitive learning scenario need to be reformulated and established.

\subsection{Online Cost-Sensitive Bagging}

The derivation of online cost-sensitive bagging algorithms is straightforward. Since the Poisson distribution parameter $\lambda=1$ corresponds to sampling with probability $\frac{1}{N}$, sampling an example with probability $\frac{C}{N}$ can be obtained by presenting the instance to the base model $k\sim Poisson(C)$ times in online bagging. Therefore, online UnderOverBagging can be derived as in Alg.~\ref{alg:onlineunderoverbagging}.

\begin{algorithm}\footnotesize
\caption{Online UnderOverBagging Algorithm}
\label{alg:onlineunderoverbagging}
  \textbf{Input:} \(S,M,C>1\)
  \begin{algorithmic}[1]
    \FOR{\(n=1,\dots,N\)}
        \FOR{\(m=1,\dots,M\)}
            \STATE a = $\frac{m}{M}$
            \STATE If $y_n=1$, $\lambda = aC$, else $\lambda = a$
            \STATE Let $k\sim Poisson(\lambda)$
            \STATE Do $k$ times \\
             \quad Train the base learner $h_m\to Y$ using $(x_n,y_n)$
        \ENDFOR
    \ENDFOR
  \end{algorithmic}
  \raggedright\textbf{Output:} \(H(x)=\mathop{\arg \max}\limits_{y\in Y}\sum\limits _{m=1}^M I(h_m(x)=y)\)
\end{algorithm}

\begin{algorithm}[b]\footnotesize
\caption{Online SMOTEBagging Algorithm}
\label{alg:onlinesmotebagging}
  \textbf{Input:} \(S,M,C>1\)
  \begin{algorithmic}[1]
    \STATE $X^+=\{\}$
    \FOR{\(n=1,\dots,N\)}
        \FOR{\(m=1,\dots,M\)}
            \STATE a = $\frac{m}{M}$
            \IF{$y_n=1$}
            \STATE $X^+=\{X^+;x_n\}$,
            \STATE $\lambda =aC, \lambda_{SMOTE}=(1-a)C$
            \STATE Let $k\sim Poisson(\lambda)$
            \STATE Do $k$ times \\
             \quad Train the base learner $h_m\to Y$ using \\ \quad $(x_n,y_n)$
            \STATE Let $k_{SMOTE}\sim Poisson(\lambda_{SMOTE})$
            \STATE Do $k_{SMOTE}$ times \\
             \quad $x_{S}=Online\_SMOTE(X^+)$ \\
             \quad Train the base learner $h_m\to Y$ using $(x_{S},y_n)$
            \ELSE
            \STATE Let $k\sim Poisson(\lambda)$
            \STATE Do $k$ times \\
             \quad Train the base learner $h_m\to Y$ using \\ \quad $(x_n,y_n)$
            \ENDIF
        \ENDFOR
    \ENDFOR
  \end{algorithmic}
  \raggedright\textbf{Output:} \(H(x)=\mathop{\arg \max}\limits_{y\in Y}\sum\limits _{m=1}^M I(h_m(x)=y)\)

\end{algorithm}

For the online SMOTEBagging, we have the similar generalization, which is shown in Alg.~\ref{alg:onlinesmotebagging}, together with online SMOTE in Alg.~\ref{alg:onlinesmote}. It is worth noting that the online SMOTE may not be a good approximate of its batch counterpart, since at the early stage of the algorithm, the positive examples are extremely rare, and therefore the generated synthetic examples by online SMOTE can be much different from the ones by batch SMOTE. This is an intrinsic problem of online SMOTE, and the distributions of synthetic examples created by online SMOTE and batch SMOTE cannot be the same unless we can obtain a large number of minority examples from the very beginning. However, even with this problem, experimental results demonstrate reasonably good consistency between online and batch SMOTEBagging algorithms.

\begin{algorithm}\footnotesize
\caption{Online SMOTE Algorithm}
\label{alg:onlinesmote}
  \textbf{Input:} \(X\)
  \begin{algorithmic}[1]
    \STATE $x=X(end,:)$
    \STATE Randomly choose one of the $k$ nearest neighbors of $x$, $x'$
    \STATE Calculate the difference between $x$ and $x'$
    \STATE Generate a random number $\gamma$ between 0 and 1
    \STATE Create a synthetic instance: \\ $x_{S} = x + \gamma (x'-x)$
  \end{algorithmic}
  \raggedright\textbf{Output:} a synthetic instances $x_{SMOTE}$
\end{algorithm}

\subsection{Online Cost-Sensitive Boosting}
The standard online boosting framework can be generalized to be cost-sensitive by introducing different parameters of Poisson distribution for different classes. As stated at the beginning of this section, to implement online cost-sensitive boosting, the key step is to formulate new weight updates formulas of the batch algorithms that do not require the normalization step.

For AdaC2 algorithm, we have
\begin{align*}
&D_{m+1}(n) \\
&\propto C_nD_m(n)exp(-\alpha_m(2I(h_m(x_n)=y_n)-1)) \\
&= C_nD_m(n)exp(-2\alpha_mI(h_m(x_n)=y_n))exp(\alpha) \\
&\propto C_nD_m(n)exp(-2\alpha_mI(h_m(x_n)=y_n)) \\
&= D_m\times
    \begin{cases}
        \frac{C_n\sum_{n=1}^NC_nD_m(n)I(h_m(x_n)\neq y_n)}{\sum_{n=1}^NC_nD_m(n)I(h_m(x_n)=y_n)},  \ h_m(x_n)=y_n\\
        C_n, \hfill h_m(x_n)\neq y_n
    \end{cases}\\
&\propto D_m\times
    \begin{cases}
        \frac{C_n}{2\sum_{n=1}^NC_nD_m(n)I(h_m(x_n)=y_n)},      \quad h_m(x_n)=y_n\\
        \frac{C_n}{2\sum_{n=1}^NC_nD_m(n)I(h_m(x_n)\neq y_n)},  \quad h_m(x_n)\neq y_n
    \end{cases}.
\end{align*}

Let $werr_m=\sum_{n=1}^NC_nD_m(n)I(h_m(x_n)\neq y_n)$ be the weighted error and $wacc=\sum_{n=1}^NC_nD_m(n)I(h_m(x_n)= y_n)$ be the weighed accuracy. If we reformulate the weight update step (step 6) of AdaC2 as
\begin{align*}
&D_{m+1}(n) = D_m\times
            \begin{cases}
                h_m(x_n)=y_n
                    \begin{cases} y_n=1, \frac{C_P}{2wacc_m}\\
                    y_n=0,  \frac{C_N}{2wacc_m}
                    \end{cases}\\
                h_m(x_n)\neq y_n
                    \begin{cases} y_n=1,  \frac{C_P}{2werr_m}\\
                    y_n=0,  \frac{C_N}{2werr_m}
                    \end{cases}
            \end{cases},
\end{align*}
it can be observed that it is equivalent to step 6 of Alg.~\ref{alg:AdaC2} but without the normalization factor, and the sum of the weights after the update is still one. Therefore, to implement online AdaC2, we only need to track $werr$ and $wacc$ to update the Poisson parameter $\lambda$ for each base learner. In particular, as AdaC2 treats differently true positive, true negative, false positive and false negative examples, we need four parameters ($\lambda _m^{TP}, \lambda _m^{TN}, \lambda _m^{FP}, \lambda _m^{FN}$) to track the sum of weighted Poisson distribution parameter $\lambda$ of each category of the examples for the $m$th base model respectively. Then $werr$ and $wacc$ can be calculated by $\frac{\lambda _m^{FP}+\lambda _m^{FN}}{\lambda _m^{SUM}}$ and $\frac{\lambda _m^{TP}+\lambda _m^{TN}}{\lambda _m^{SUM}}$ respectively, where $\lambda _m^{SUM}$ is the unweighted sum of total Poisson distribution parameter for the $m$th base learner. The pseudo-code of online AdaC2 is shown in Alg.~\ref{alg:onlineAdaC2}.

\begin{algorithm}\footnotesize
\caption{Online AdaC2 Algorithm}
\label{alg:onlineAdaC2}
  \textbf{Input:} \(S,M,C_N,C_P\)
  \begin{algorithmic}[1]
  \STATE Initialize $\lambda_m^{TP} = 0, \lambda_m^{TN} = 0, \lambda_m^{FP} = 0, \lambda_m^{FN} = 0,$ $\lambda_m^{SUM} = 0$ for all $m\in \{1,\dots,M\}$
    \FOR{\(n=1,\dots,N\)}
        \STATE Set $\lambda = 1$
        \FOR{\(m=1,\dots,M\)}
            \STATE $\lambda_m^{SUM} = \lambda_m^{SUM}+\lambda$
            \STATE Let $k\sim Poisson(\lambda)$
            \STATE Do $k$ times
            \STATE \quad Train the base learner $h_m\to Y$ using $(x_n,y_n)$
            \IF{$h_m(x_n)=1\&\&y_n=1$}
                \STATE $\lambda_m^{TP} \leftarrow \lambda_m^{TP} +C_P\lambda,$ \\
                $wacc_m\leftarrow \frac{\lambda_m^{TP}+\lambda_m^{TN}}{\lambda_m^{SUM}},
                werr_m\leftarrow \frac{\lambda_m^{FP}+\lambda_m^{FN}}{\lambda_m^{SUM}},$ \\
                $\lambda \leftarrow \frac{C_P\lambda}{2wacc_m}$
            \ELSIF{$h_m(x_n)=0\&\&y_n=0$}
                \STATE $\lambda_m^{TN} \leftarrow \lambda_m^{TN} +C_N\lambda,$ \\
                $wacc_m\leftarrow \frac{\lambda_m^{TP}+\lambda_m^{TN}}{\lambda_m^{SUM}},
                werr_m\leftarrow \frac{\lambda_m^{FP}+\lambda_m^{FN}}{\lambda_m^{SUM}},$ \\
                $\lambda \leftarrow \frac{C_N\lambda}{2wacc_m}$
            \ELSIF{$h_m(x_n)=0\&\&y_n=1$}
                \STATE $\lambda_m^{FN} \leftarrow \lambda_m^{FN} +C_P\lambda,$ \\
                $wacc_m\leftarrow \frac{\lambda_m^{TP}+\lambda_m^{TN}}{\lambda_m^{SUM}},
                werr_m\leftarrow \frac{\lambda_m^{FP}+\lambda_m^{FN}}{\lambda_m^{SUM}},$ \\
                $\lambda \leftarrow \frac{C_P\lambda}{2werr_m}$
            \ELSIF{$h_m(x_n)=1\&\&y_n=0$}
                \STATE $\lambda_m^{FP} \leftarrow \lambda_m^{FP} +C_N\lambda,$ \\
                $wacc_m\leftarrow \frac{\lambda_m^{TP}+\lambda_m^{TN}}{\lambda_m^{SUM}},
                werr_m\leftarrow \frac{\lambda_m^{FP}+\lambda_m^{FN}}{\lambda_m^{SUM}},$ \\
                $\lambda \leftarrow \frac{C_N\lambda}{2werr_m}$
            \ENDIF
        \ENDFOR
    \ENDFOR
  \end{algorithmic}
  \raggedright\textbf{Output:} \(H(x)=\mathop{\arg \max}\limits_{y\in Y}\sum\limits _{m=1}^M log(\frac{wacc_m}{werr_m})I(h_m(x)=y)\)
\end{algorithm}

By similar calculation, the update formula of CSB2 without normalization can be formulated as
\begin{align*}
&D_{m+1}(n) = D_m\times
    \begin{cases}
        h_m(x_n)=y_n
            \quad \frac{\epsilon_m}{(1-\epsilon_m)(\epsilon_m+werr_m)}\\
        h_m(x_n)\neq y_n
            \begin{cases} y_n=1,  \hfill\quad\frac{C_P}{\epsilon_m+werr_m}\\
            y_n=0,  \hfill\quad \frac{C_N}{\epsilon_m+werr_m}
            \end{cases}.
    \end{cases},
\end{align*}
Then, the online CSB2 can be implemented by tracking the weighted error and unweighted error, which can be done by tracking $\lambda _m^{FP}, \lambda _m^{FN}$, $\lambda _m^{SW}$ and $\lambda _m^{SUM}$ respectively. The pseudo-code of online CSB2 is shown in Alg.~\ref{alg:onlineCSB2}.
\begin{algorithm}[h]\footnotesize
\caption{Online CSB2 Algorithm}
\label{alg:onlineCSB2}
  \textbf{Input:} \(S,M,C_N,C_P\)
  \begin{algorithmic}[1]
  \STATE Initialize $\lambda_m^{FP}=0, \lambda_m^{FN}=0, \lambda_m^{SW}=0, \lambda_m^{SUM}=0$ for all $m\in \{1,\dots,M\}$
    \FOR{\(n=1,\dots,N\)}
        \STATE Set $\lambda = 1$
        \FOR{\(m=1,\dots,M\)}
            \STATE $\lambda_m^{SUM} = \lambda_m^{SUM}+\lambda$
            \STATE Let $k\sim Poisson(\lambda)$
            \STATE Do $k$ times
            \STATE \quad Train the base learner $h_m\to Y$ using $(x_n,y_n)$
            \IF{$h_m(x_n)=y_n$}
                \STATE $\epsilon_m\leftarrow \frac{\lambda_m^{SW}}{\lambda_m^{SUM}},
                werr_m\leftarrow \frac{\lambda_m^{FP}+\lambda_m^{FN}}{\lambda_m^{SUM}},$ \\
                $\lambda \leftarrow \frac{\epsilon_m}{(1-\epsilon_m)(\epsilon_m+werr_m)}$
            \ELSE
                \IF{$h_m(x_n)=0\&\&y_n=1$}
                    \STATE $\lambda_m^{FN} \leftarrow \lambda_m^{FN}+C_P\lambda,\lambda_m^{SW} \leftarrow \lambda_m^{SW} +\lambda,$ \\
                    $\epsilon_m\leftarrow \frac{\lambda_m^{SW}}{\lambda_m^{SUM}},
                    werr_m\leftarrow \frac{\lambda_m^{FP}+\lambda_m^{FN}}{\lambda_m^{SUM}},$ \\
                    $\lambda \leftarrow \frac{C_P\lambda}{\epsilon_m+werr_m}$
                \ELSE
                    \STATE $\lambda_m^{FP} \leftarrow \lambda_m^{FP}+C_P\lambda,\lambda_m^{SW} \leftarrow \lambda_m^{SW} +\lambda,$ \\
                    $\epsilon_m\leftarrow \frac{\lambda_m^{SW}}{\lambda_m^{SUM}},
                    werr_m\leftarrow \frac{\lambda_m^{FP}+\lambda_m^{FN}}{\lambda_m^{SUM}},$ \\
                    $\lambda \leftarrow \frac{C_N\lambda}{\epsilon_m+werr_m}$
                \ENDIF
            \ENDIF
        \ENDFOR
    \ENDFOR
  \end{algorithmic}
  \raggedright\textbf{Output:} \(H(x)=\mathop{\arg \max}\limits_{y\in Y}\sum\limits _{m=1}^M log(\frac{1-\epsilon_m}{\epsilon_m})I(h_m(x)=y)\)
\end{algorithm}

As shown in Alg.~\ref{alg:rusboost}, the RUSBoost algorithm is the same as the standard boosting algorithm except that there is a preprocessing step at the beginning of every iteration to rebalance the example distribution. As a result, each example should be reweighted according to different undersampling strategies. In RUSBoost1, as all negative examples are first undersampled at a rate of $\frac{CN^+}{N^-}$, the expected total weights of the remaining examples are $D_m^++\frac{CN^+}{N^-}D_m^-$, where $D_m^+=\sum_{n=1}^{N^+}D_m^+(n)$ is the sum of the weights of positive examples, and $D_m^-=\sum_{n=1}^{N^-}D_m^-(n)$ is the one of negative examples. Since the size of training examples is $(C+1)N^+$ after undersampling, we finally get the new weight of each example:
\begin{equation*}
  D_m'(n)=D_m(n)\times
  \begin{cases} \frac{1}{D_m^++\frac{CN^+}{N^-}D_m^-}\frac{(C+1)N^+}{N^++N^-}, \hfill y_n=1\\
                \frac{CN^+}{N^-}\frac{1}{D_m^++\frac{CN^+}{N^-}D_m^-}\frac{(C+1)N^+}{N^++N^-}, y_n=0
  \end{cases}\\.
\end{equation*}
Therefore, to approximate these quantities in online RUSBoost1, we need to record the total weights of positive examples and negative examples, as well as the total number of them respectively, which can be done by four additional parameters: $\lambda_m^{POS}$, $\lambda_m^{NEG}$, $n^+$, and $n^-$. In particular, $N^+$ and $N^-$ can be approximated by $n^+$, and $n^-$. Since $D_m(n)$ in batch boosting (Alg.~\ref{alg:adaboost}) corresponds to the parameter $\lambda$ in online boosting (Alg.~\ref{alg:onlineadaboost}), $D_m^+$ and $D_m^-$ can be tracked by $\frac{\lambda_m^{POS}}{\lambda_m^{POS}+\lambda_m^{NEG}}$ and $\frac{\lambda_m^{NEG}}{\lambda_m^{POS}+\lambda_m^{NEG}}$ respectively. The derivation of online RUSBoost2 follows a similar approach. In RUSBoost2, the number of examples of each class is fixed regardless of the total weight of each class. Therefore, the positive (negative) examples will be undersampled if the total weight of them is larger than $\frac{N^+}{N^++N^-}$($\frac{CN^+}{N^++N^-}$), and will be oversampled otherwise. Therefore, each example is reweighted by
\begin{equation*}
  D_m'(n)=D_m(n)\times
  \begin{cases} \frac{N^+}{N^++N^-}/D_m^+,  \quad y_n=1\\
                \frac{CN^+}{N^++N^-}/D_m^-, \quad y_n=0
  \end{cases}\\.
\end{equation*}
The implementation of online RUSBoost3 is straightforward, positive examples are generated as standard online boosting. For a negative example, let $\lambda^{RUS}=\frac{\lambda}{C}$, and use this example $k\sim Poisson(\lambda^{RUS})$ times to update a base learner. The pseudo-code of these three online RUSBoost algorithms is demonstrated in Alg.~\ref{alg:onlineRUS}.

\begin{algorithm}\footnotesize
\caption{Online RUSBoost Algorithm}
\label{alg:onlineRUS}
  \textbf{Input:} \(S,M,C\)
  \begin{algorithmic}[1]
  \STATE Initialize $\lambda_m^{SC} = 0, \lambda_m^{SW} = 0, \lambda_m^{POS} = 0, \lambda_m^{NEG} = 0,$ for all $m\in \{1,\dots,M\}$, $n^+=0$, $n^-=0$
    \FOR{\(n=1,\dots,N\)}
        \STATE Set $\lambda = 1$
        \FOR{\(m=1,\dots,M\)}
            \IF{$y_n=1$}
                \STATE $\lambda_m^{POS} \leftarrow \lambda_m^{POS} +\lambda,$ $n^+ \leftarrow n^++1,$\\
            \ELSE
                \STATE $\lambda_m^{NEG} \leftarrow \lambda_m^{NEG} +\lambda,$ $n^- \leftarrow n^-+1,$\\
            \ENDIF
            \STATE \underline{Compute $\lambda^{RUS}$} (step 21 -- 23)
            \STATE Let $k\sim Poisson(\lambda^{RUS})$
            \STATE Do $k$ times
            \STATE \quad Train the base learner $h_m\to Y$ using $(x_n,y_n)$
            \IF{$h_m(x_n)=y_n$}
                \STATE $\lambda_m^{SC} \leftarrow \lambda_m^{SC} +\lambda,\epsilon_m\leftarrow \frac{\lambda_m^{SW}}{\lambda_m^{SC} + \lambda_m^{SW}}, \lambda \leftarrow \frac{\lambda}{2(1-\epsilon_m)}$
            \ELSE
                \STATE $\lambda_m^{SW} \leftarrow \lambda_m^{SW} +\lambda,\epsilon_m\leftarrow \frac{\lambda_m^{SW}}{\lambda_m^{SC} + \lambda_m^{SW}}, \lambda \leftarrow \frac{\lambda}{2\epsilon_m}$
            \ENDIF
        \ENDFOR
    \ENDFOR
  \raggedright\textbf{Output:} \(H(x)=\mathop{\arg \max}\limits_{y\in Y}\sum\limits _{m=1}^M log(\frac{1-\epsilon_m}{\epsilon_m})I(h_m(x)=y)\)
  \rule{\linewidth}{.1pt}
  \textbf{online RUSBoost1}  \\
  \STATE Compute $\lambda^{RUS}$:
            $\begin{cases}
                \lambda^{RUS} \leftarrow \lambda\frac{\lambda_m^{POS}+\lambda_m^{NEG}}{\lambda_m^{POS}+\lambda_m^{NEG}\frac{Cn^+}{n^-}}\frac{(C+1)n^+}{n^++n^-},\quad y_n=1 \\
                \lambda^{RUS} \leftarrow \lambda\frac{\lambda_m^{POS}+\lambda_m^{NEG}}{\lambda_m^{NEG}+\lambda_m^{POS}\frac{n^-}{Cn^+}}\frac{(C+1)n^+}{n^++n^-},\hfill y_n=0
            \end{cases} $
  \rule{\linewidth}{.1pt}
  \textbf{online RUSBoost2}  \\
  \STATE Compute $\lambda^{RUS}$:
            $\begin{cases}
                \lambda^{RUS} \leftarrow \lambda\frac{n^+}{n^++n^-}/\frac{\lambda^{POS}}{\lambda^{POS}+\lambda^{NEG}},\quad y_n=1 \\
                \lambda^{RUS} \leftarrow \lambda\frac{Cn^+}{n^++n^-}/\frac{\lambda^{NEG}}{\lambda^{POS}+\lambda^{NEG}},\hfill y_n=0
            \end{cases} $
  \rule{\linewidth}{.1pt}
  \textbf{online RUSBoost3}  \\
  \STATE Compute $\lambda^{RUS}$:
            $\begin{cases}
                \lambda^{RUS} \leftarrow \lambda,\quad y_n=1 \\
                \lambda^{RUS} \leftarrow \frac{\lambda}{C},\hfill y_n=0
            \end{cases} $
  \end{algorithmic}
\end{algorithm}

The pseudo-code of the online SMOTEBoost algorithms are described in Alg.~\ref{alg:onlineSMOTE}. In online SMOTEBoost1, the weights of synthetic examples are $\frac{1}{N'}$, where $N'$ is the number of examples in the new data set. As a result, the probability that an example in the original data set is sampled is the same as its weight before renormalization, which means the parameter $\lambda$ used in online SMOTEBoost1 is the same as the standard online boosting. Besides, synthetic examples are generated by uniformly sampling from positive examples and applying SMOTE. In online SMOTEBoost2, the derivation of $\lambda'$ is the same as $\lambda^{RUS}$ in online RUSBoost2. Since SMOTE is applied after sampling the original data set, the probability of a positive example being chosen to generate synthetic examples is proportional to its weight. Therefore, the parameter $\lambda^{SMOTE}$ is given by $\lambda'(\frac{N^-}{CN^+}-1)$. The online SMOTEBoost3 is the same as standard online boosting, except for the additional parameter $\lambda^{SMOTE}=\lambda(C-1)$.

\begin{algorithm}\footnotesize
\caption{Online SMOTEBoost Algorithm}
\label{alg:onlineSMOTE}
  \textbf{Input:} \(S,M,C\)
  \begin{algorithmic}[1]
  \STATE Initialize $\lambda_m^{SC} = 0, \lambda_m^{SW} = 0,$ for all $m\in \{1,\dots,M\}$, $n^+=0$, $n^-=0$, $X^+=\{\}$
    \FOR{\(n=1,\dots,N\)}
        \STATE Set $\lambda = 1$
        \FOR{\(m=1,\dots,M\)}
            \IF{$y_n=1$}
                \STATE $n^+ \leftarrow n^++1$, $\lambda_m^{POS} \leftarrow \lambda_m^{POS}+\lambda$,\\
                 $X^+=\{X^+;x_n\}$ \\
            \ELSE
                \STATE $n^- \leftarrow n^-+1$, $\lambda_m^{NEG} \leftarrow \lambda_m^{NEG}+\lambda$,
            \ENDIF
            \STATE \underline{Compute $\lambda'$} (step 26, 28, 30)
            \STATE $k\sim Poisson(\lambda')$
            \STATE Do $k$ times
            \STATE \quad Train the base learner $h_m\to Y$ using $(x_n,y_n)$
            \STATE \underline{Compute $\lambda^{SMOTE}$} (step 27, 29, 31)
            \STATE $k^{SMOTE}\sim Poisson(\lambda^{SMOTE})$\\
            \STATE Do $k^{SMOTE}$ times
            \STATE \quad $x_{S}=online\_SMOTE(X^+)$
            \STATE \quad Train the base learner $h_m\to Y$ using $(x_{S},y_n)$

            \IF{$h_m(x_n)=y_n$}
                \STATE $\lambda_m^{SC} \leftarrow \lambda_m^{SC} +\lambda,\epsilon_m\leftarrow \frac{\lambda_m^{SW}}{\lambda_m^{SC} + \lambda_m^{SW}}, \lambda \leftarrow \frac{\lambda}{2(1-\epsilon_m)}$
            \ELSE
                \STATE $\lambda_m^{SW} \leftarrow \lambda_m^{SW} +\lambda,\epsilon_m\leftarrow \frac{\lambda_m^{SW}}{\lambda_m^{SC} + \lambda_m^{SW}}, \lambda \leftarrow \frac{\lambda}{2\epsilon_m}$
            \ENDIF
        \ENDFOR
    \ENDFOR
  \raggedright\textbf{Output:} \(H(x)=\mathop{\arg \max}\limits_{y\in Y}\sum\limits _{m=1}^M log(\frac{1-\epsilon_m}{\epsilon_m})I(h_m(x)=y)\)
  \rule{\linewidth}{.1pt}
  \textbf{online SMOTEBoost1}\\
  \STATE Compute $\lambda'$: $\lambda' \leftarrow \lambda$ \\
  \STATE Compute $\lambda^{SMOTE}$:
            $\begin{cases}
                \lambda^{SMOTE} \leftarrow \frac{n^-}{Cn^+}-1, \quad y_n=1\\
                \lambda^{SMOTE} \leftarrow 0,\hfill y_n=0
            \end{cases} $
  \rule{\linewidth}{.1pt}
  \textbf{online SMOTEBoost2}\\
  \STATE Compute $\lambda'$:
            $\begin{cases}
                \lambda' \leftarrow \lambda\frac{n^+}{n^++n^-}/\frac{\lambda^{POS}}{\lambda^{POS}+\lambda^{NEG}}, \quad y_n=1\\
                \lambda' \leftarrow \lambda\frac{n^-}{n^++n^-}/\frac{\lambda^{NEG}}{\lambda^{POS}+\lambda^{NEG}},\hfill y_n=0
            \end{cases} $
  \STATE Compute $\lambda^{SMOTE}$:
            $\begin{cases}
                \lambda^{SMOTE} \leftarrow \lambda'(\frac{n^-}{Cn^+}-1), \ y_n=1\\
                \lambda^{SMOTE} \leftarrow 0,\hfill y_n=0
            \end{cases} $
            \rule{\linewidth}{.1pt}
  \textbf{online SMOTEBoost3}\\
  \STATE Compute $\lambda'$: $\lambda' \leftarrow \lambda$ \\
  \STATE Compute $\lambda^{SMOTE}$:
            $\begin{cases}
                \lambda^{SMOTE} \leftarrow (C-1)\lambda, \quad y_n=1\\
                \lambda^{SMOTE} \leftarrow 0,\hfill y_n=0
            \end{cases} $
\end{algorithmic}
\end{algorithm}

\section{Theoretical Analysis}

In this section, we prove that under the similar conditions as for cost-insensitive online bagging and boosting \cite{oza2001online}, the online UnderOverBagging, AdaC2, CSB2 and RUSBoost converge to the same solution as their batch counterparts. SMOTEBagging and SMOTEBoost theoretically cannot converge to their batch mode counterparts since online SMOTE cannot converge to batch SMOTE. Because of the space limitation, we only give the statements of the main theorems.

\newtheorem{theory}{Theorem}

\begin{theory}
As $N^+\to \infty$, $N^-\to \infty$ and $M\to \infty$, if the base learning algorithms are proportional, and converge in probability to some classifier, online UnderOverBagging converges to its batch mode counterpart.
\end{theory}




\begin{theory}
As $N^+\to \infty$ and $N^-\to \infty$, if the base learners are naive Bayes classifiers, online AdaC2 and CSB2 converge to their batch mode counterparts.
\end{theory}


\begin{theory}
As $N^+\to \infty$ and $N^-\to \infty$, if the base learners are naive Bayes classifiers, the online RUSBoost algorithms converge to their batch mode counterparts.
\end{theory}

The proofs of the theorems follows quite readily from the work by \cite{oza2001onlinephd}, but need to be generalized to different cost settings. The details of the proofs are in the Supplementary Materials.

\section{Experiments}
In this section, the proposed online cost-sensitive ensemble learning algorithms were compared with their batch mode counterparts using benchmark data sets, and the performances, measured by area under an ROC curve (AUC), were obtained by five-fold cross-validation. Since the performance of different batch ensemble algorithms under different conditions has been thoroughly compared in \cite{galar2012review}, \cite{khoshgoftaar2011comparing}, \cite{sun2007cost}, \cite{seiffert2010rusboost}, this paper mainly focuses on the performances of online algorithms, and the comparison between the two different types of learning approaches. We aim to answer the following questions:
\begin{enumerate}
\item [1.] Whether the relative performance of online cost-sensitive ensemble algorithms are consistent with that of their batch counterparts, and which ones achieve the best performance.
\item [2.] For each algorithm, how close is the performance of the online algorithm to its batch counterpart?
\item [3.] Whether the choice of base learners affects the performance and convergence of the online ensemble algorithms.
\item [4.] Though the asymptotic properties of some online ensemble algorithms have been proven, what can we observe empirically about convergence speed for the batch versus online algorithms?
\end{enumerate}

\subsection{Base Learners and Parameters}
 As the purpose of the experiments is to make fair comparisons between the proposed online algorithms and their batch mode counterparts, the effects of base learners should be reduced as much as possible. In Lemma 1, it is required that the base learners should be \emph{lossless}, \emph{proportional}, as well as \emph{stable}, which means that given the same replicas or proportion of training examples, the base learners returned by batch and online learning algorithms should be the same, and the small changes in the distribution of the examples should not cause large changes in the base learners. In addition, the algorithms should be easy to implement and the computation cost of an update step should be low in the online learning scenario. Based on these considerations, three classifiers are used in the experiments: linear discriminant analysis (LDA), quadratic discriminant analysis (QDA), and naive Bayes (NB) with Gaussian distribution for each feature. For all of the base learners, the parameters can be estimated incrementally and losslessly without storing the data stream [50]. Therefore, the differences of the performances of online and batch algorithms only come from the ensemble part.

 There are some parameters that need to be specified. In particular, we use ten base learners for all algorithms as suggested in \cite{galar2012review}. The other parameters of the ensemble algorithms are summarized in Table~\ref{tab:parameter}.

\begin{table}
\setlength{\belowcaptionskip}{10pt}
\caption{Parameter Setting}
\label{tab:parameter}
\centering
\begin{tabular}{ll}
\hline
Algorithms & Parameters \\
\hline
SMOTE                 & Number of neighbors $k=5$            \\
                      & Distance = Euclidian distance        \\
\hline
AC2, CSB2             & $C_P=1$                              \\
                      & $C_N=0.1$ to $1$                     \\
\hline
UOB, SB,RUS3,SBO3     & $C=1$ to original class ratio        \\
\hline
RUS1,RUS2,SBO1,SBO2   & $C=$ original class ratio to $1$     \\
\hline
\end{tabular}
\end{table}

\subsection{Data Sets}

Eighteen data sets from the UCI repository\footnote{Download form http://sci2s.ugr.es/keel/datasets.php.} with different class ratios were selected to compare the performances of batch and online ensemble learning algorithms. Multiclass data sets were converted to binary class data sets by selecting one class to be the positive and the rest of the examples to be the negative. Table~\ref{tab:dataset}. summarizes the characteristics of the data sets, including the number of examples (\#ex), number of features (\#fea), the percentage of positive and negative examples (\%pos; \%neg), and the class ratio (CR).

\begin{table}
\setlength{\belowcaptionskip}{10pt}
\caption{Summary of Data Sets Used in Experiments}
\label{tab:dataset}
\centering
\begin{tabular}{lllll}
\hline
Data Set & \#ex & \#fea & \%pos; \%neg & CR\\
\hline
Sonar    & 208  & 60    & 46.63, 53.37 & 1.14 \\
Glass1   & 214  & 9     & 35.51, 64.49 & 1.82 \\
Pima     & 768  & 8     & 34.84, 66.16 & 1.90 \\
Iris0    & 150  & 4     & 33.33, 66.67 & 2.00 \\
Glass0   & 214  & 9     & 32.71, 67.29 & 2.06 \\
Ecoli1   & 336  & 7     & 22.92, 77.08 & 3.36 \\
Ecoli2   & 336  & 7     & 15.48, 84.52 & 5.46 \\
Segment0 & 2308 & 19    & 14.26, 85.74 & 6.01 \\
Glass6   & 214  & 9     & 13.55, 86.45 & 6.38 \\
Yeast3   & 1484 & 8     & 10.98, 89.02 & 8.11 \\
Elico3   & 336  & 7     & 10.88, 89.12 & 8.19 \\
Satimage & 6435 & 36    & 9.73,  90.27 & 9.28 \\
Led7digit& 443  & 7     & 8.35,  91.65 & 10.97\\
Ecoli4   & 336  & 7     & 6.74,  93.26 & 13.84\\
Glass4   & 214  & 9     & 6.07,  93.93 & 15.47\\
Glass5   & 214  & 9     & 4.20,  95.80 & 22.81\\
Yeast5   & 1484 & 8     & 2.96,  97.04 & 32.78\\
Yeast6   & 1484 & 8     & 2.49,  97.51 & 39.15\\
\hline
\end{tabular}
\end{table}

\subsection{Results}

\begin{figure}[!htb]
\centering
\includegraphics[width=3.2in]{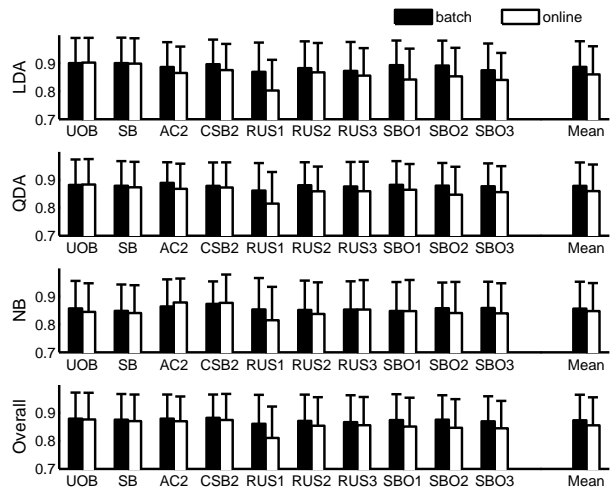}
\caption{Overall Performances with standard deviation}
\label{fig:meanDiff}
\end{figure}

Fig.~\ref{fig:meanDiff} demonstrates the average AUC with standard deviation for each algorithm and each base learner. At first glance, there is little difference between the different batch ensemble algorithms. Bagging algorithms achieve slightly better performance than boosting based approaches, which is consistent with the conclusions in [29]. On the other hand, the performance of online algorithms are much different. Roughly speaking the online algorithms can be divided into three groups. oUOB, oSB, oAC2, and oCSB2 with AUC larger than 0.87, oRUS2, oRUS3, oSBO1, oSBO2, oSBO3, with AUC between 0.84 and 0.87, and oRUS1 with AUC of 0.81.


\begin{figure}
\centering
\includegraphics[width=3.4in]{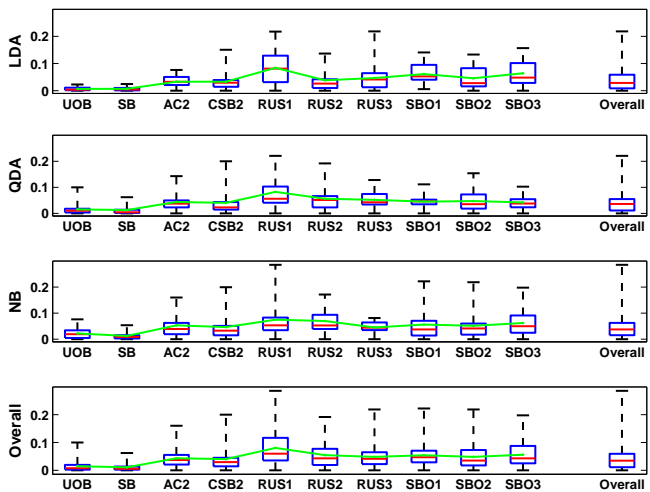}
\caption{Box-plot of absolute difference of AUC between batch and online ensemble algorithms}
\label{fig:absDiff}
\end{figure}

\begin{figure*}
\centering
\subfigure{\includegraphics[width=1.7in]{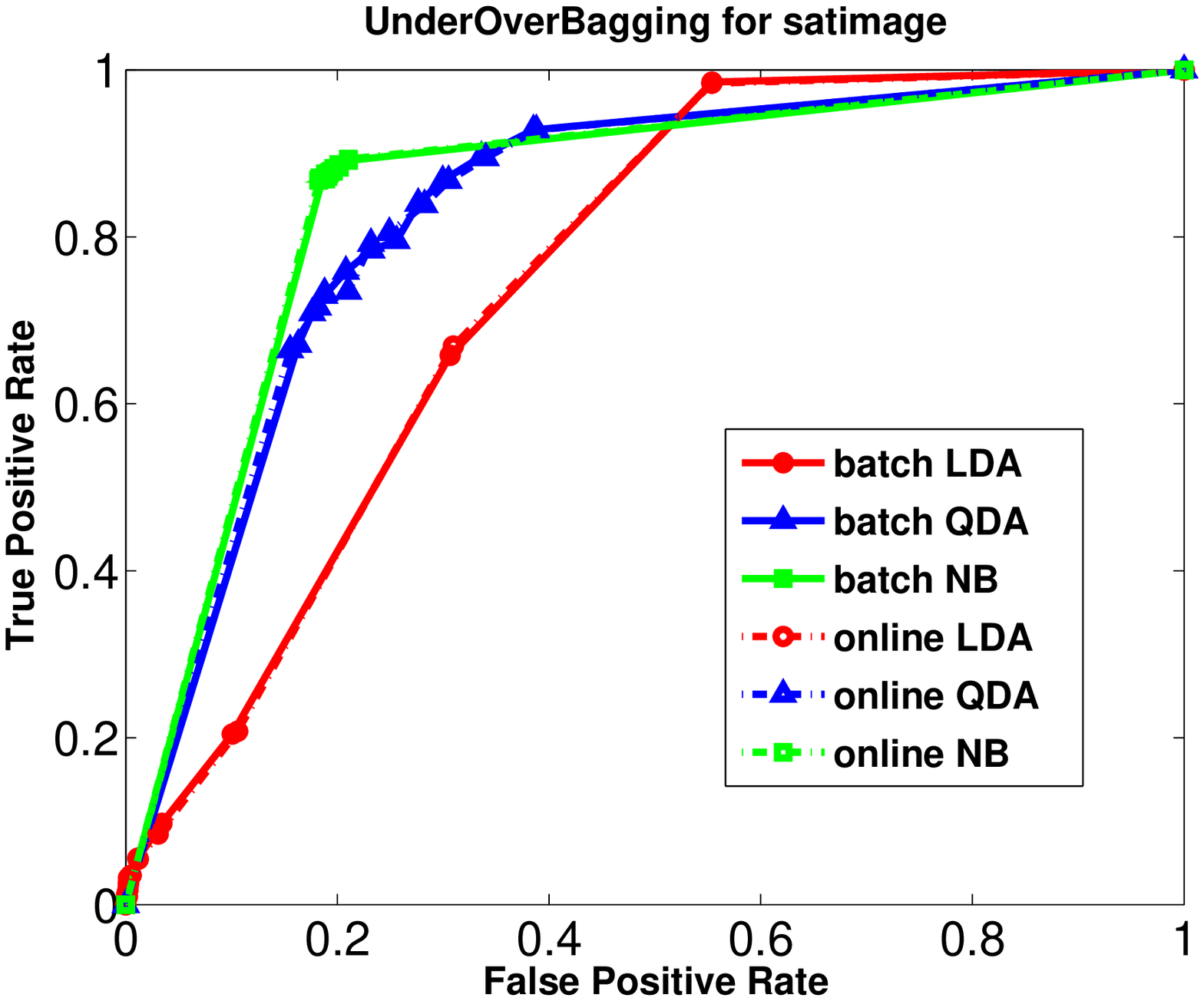}}
\subfigure{\includegraphics[width=1.7in]{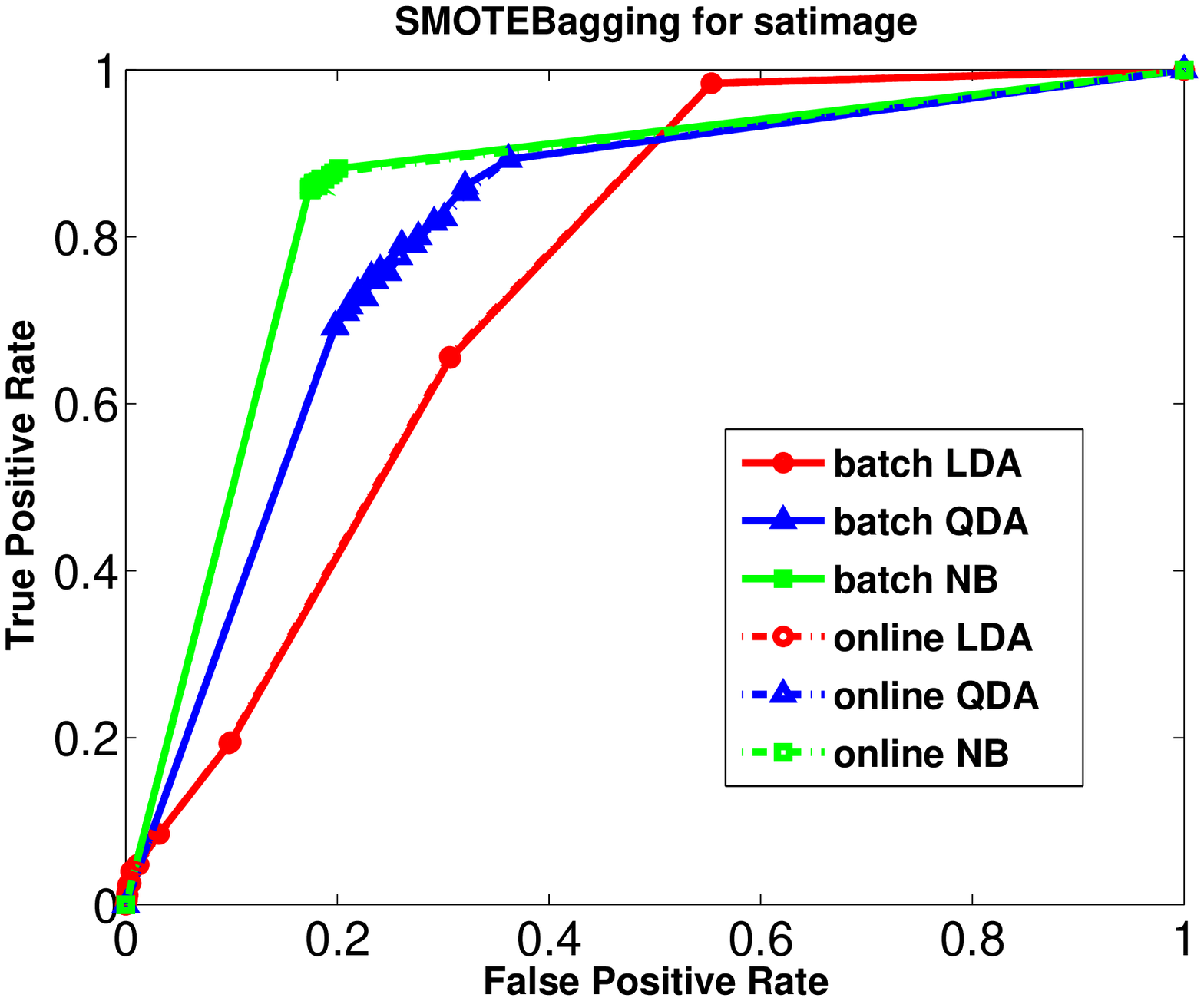}}
\subfigure{\includegraphics[width=1.7in]{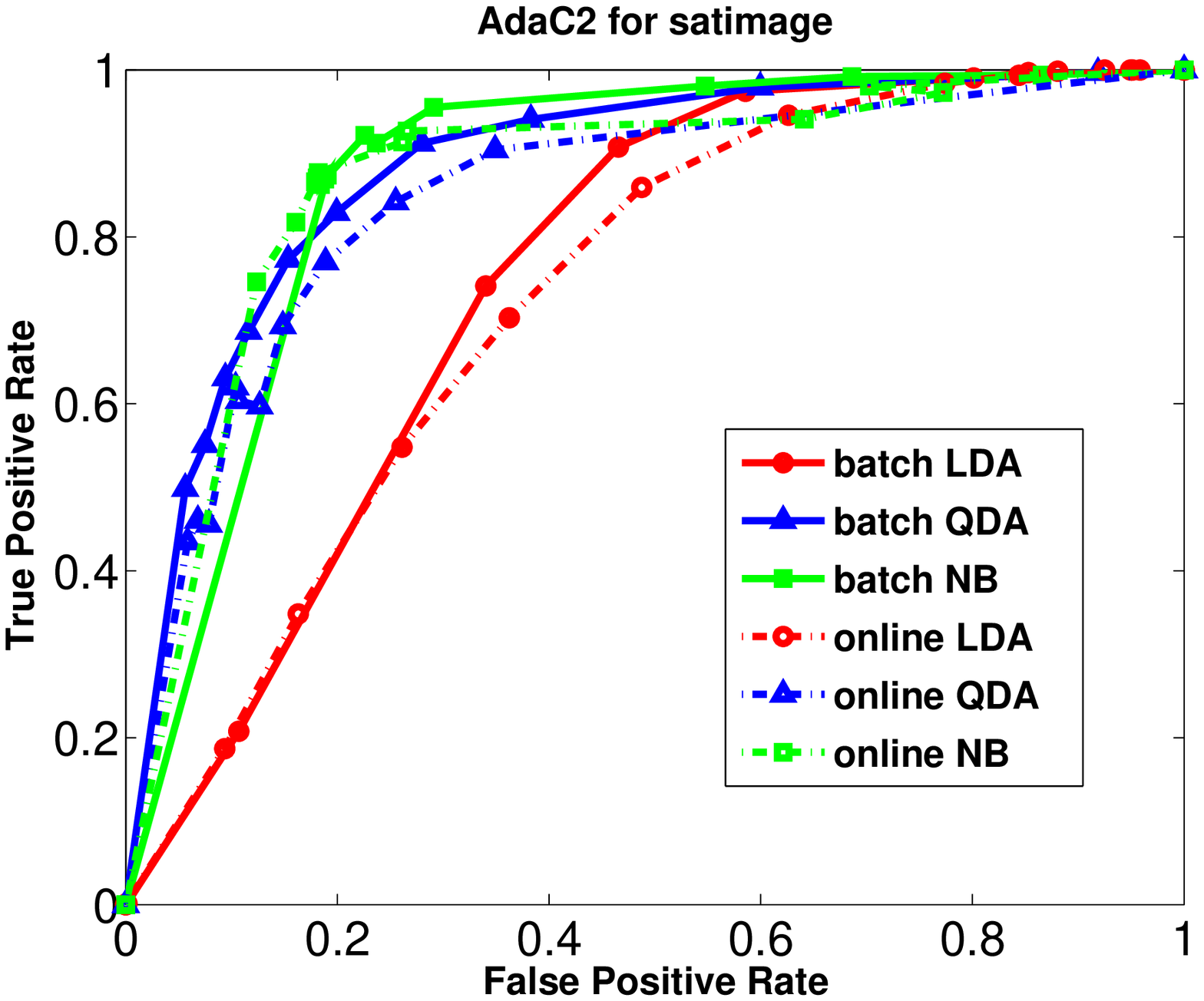}}
\subfigure{\includegraphics[width=1.7in]{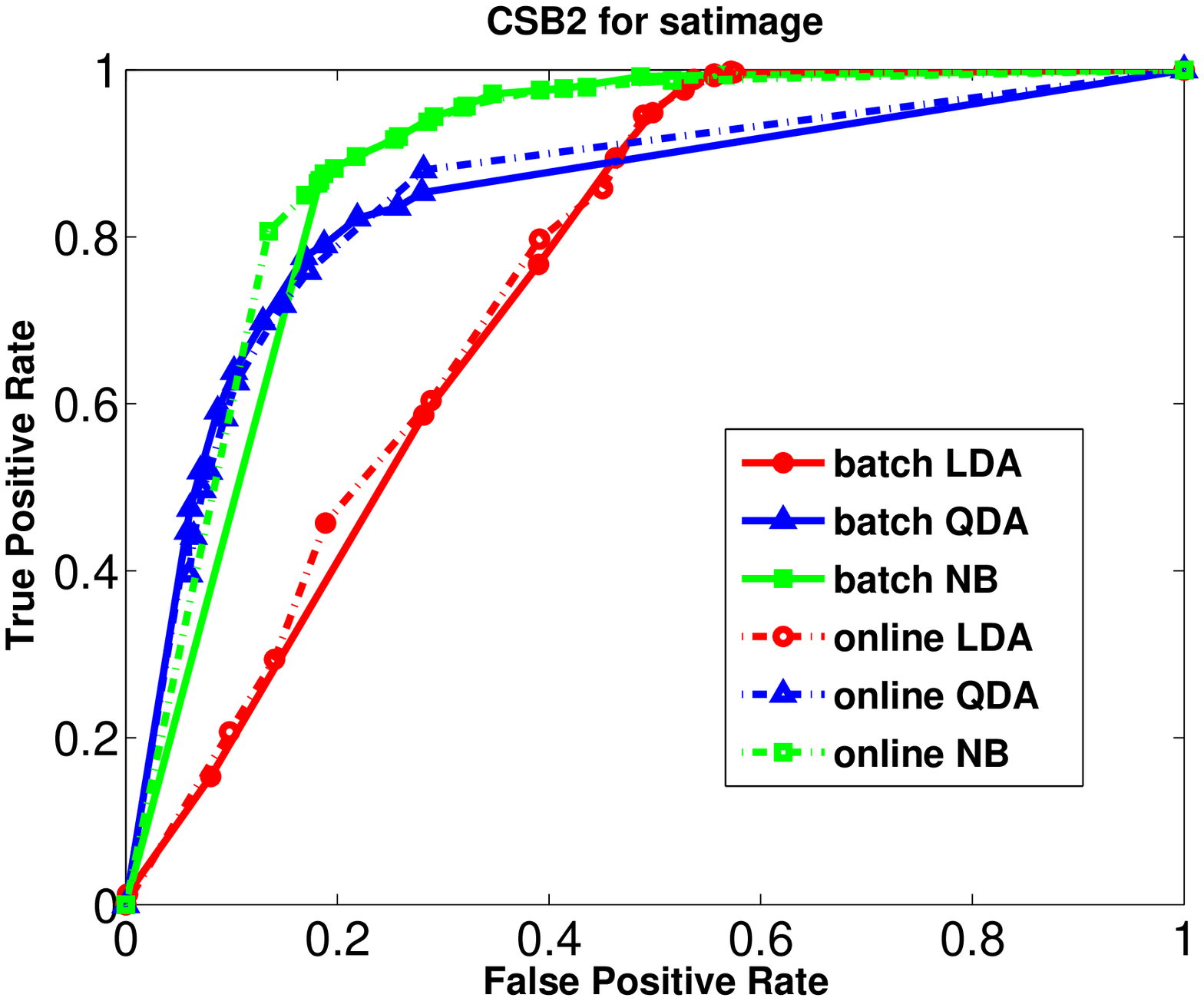}}

\subfigure{\includegraphics[width=2in]{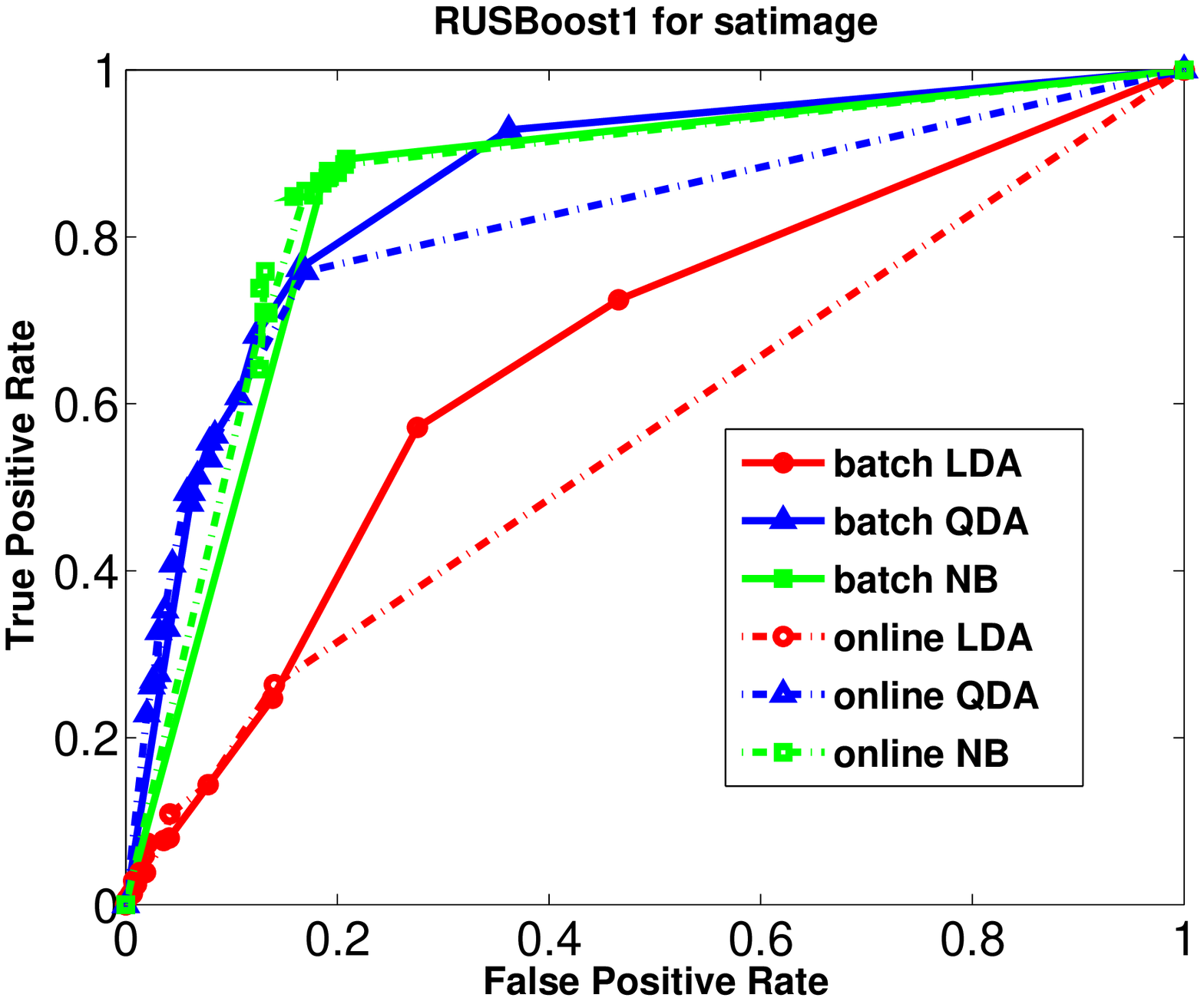}}
\hfil
\subfigure{\includegraphics[width=2in]{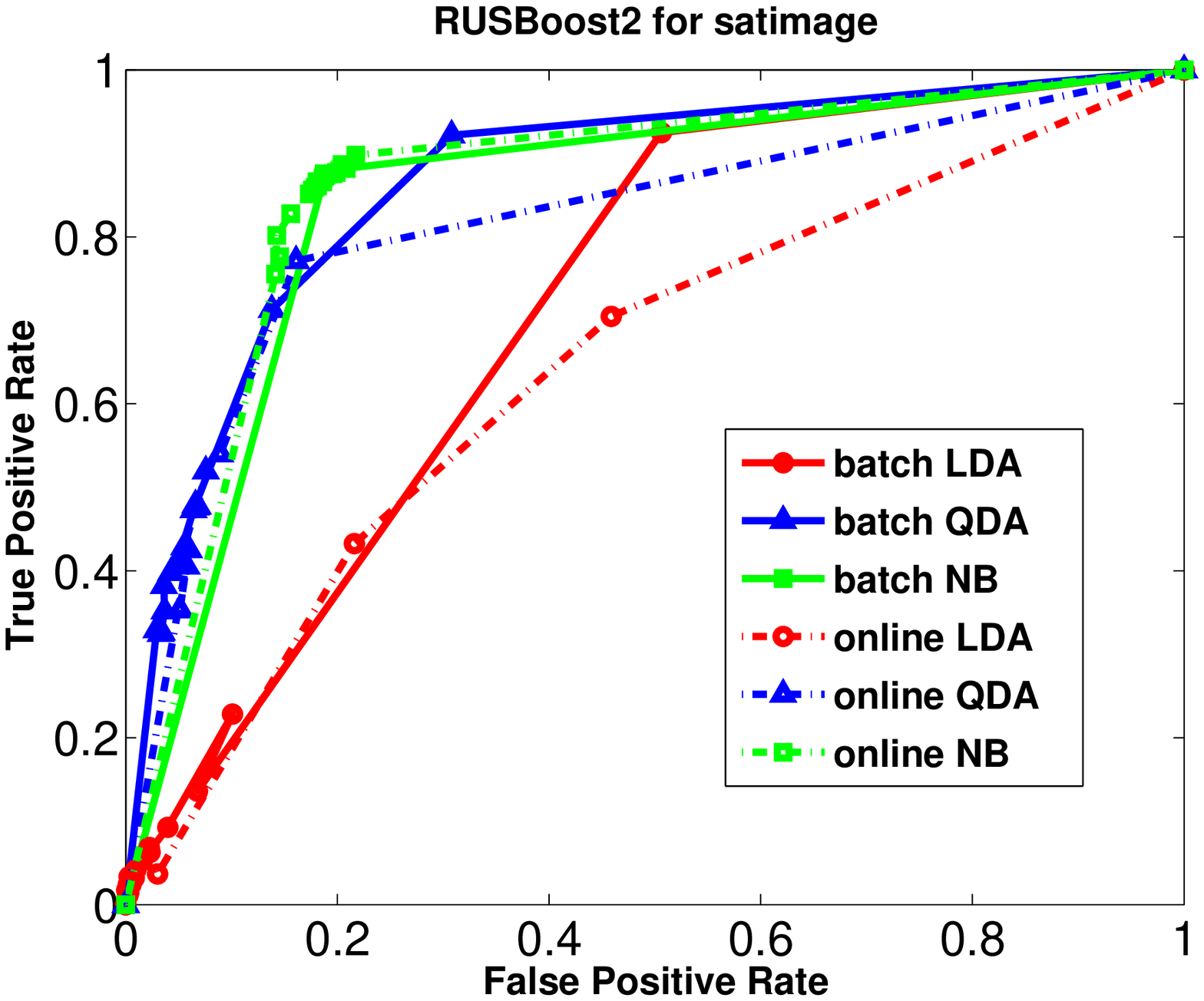}}
\hfil
\subfigure{\includegraphics[width=2in]{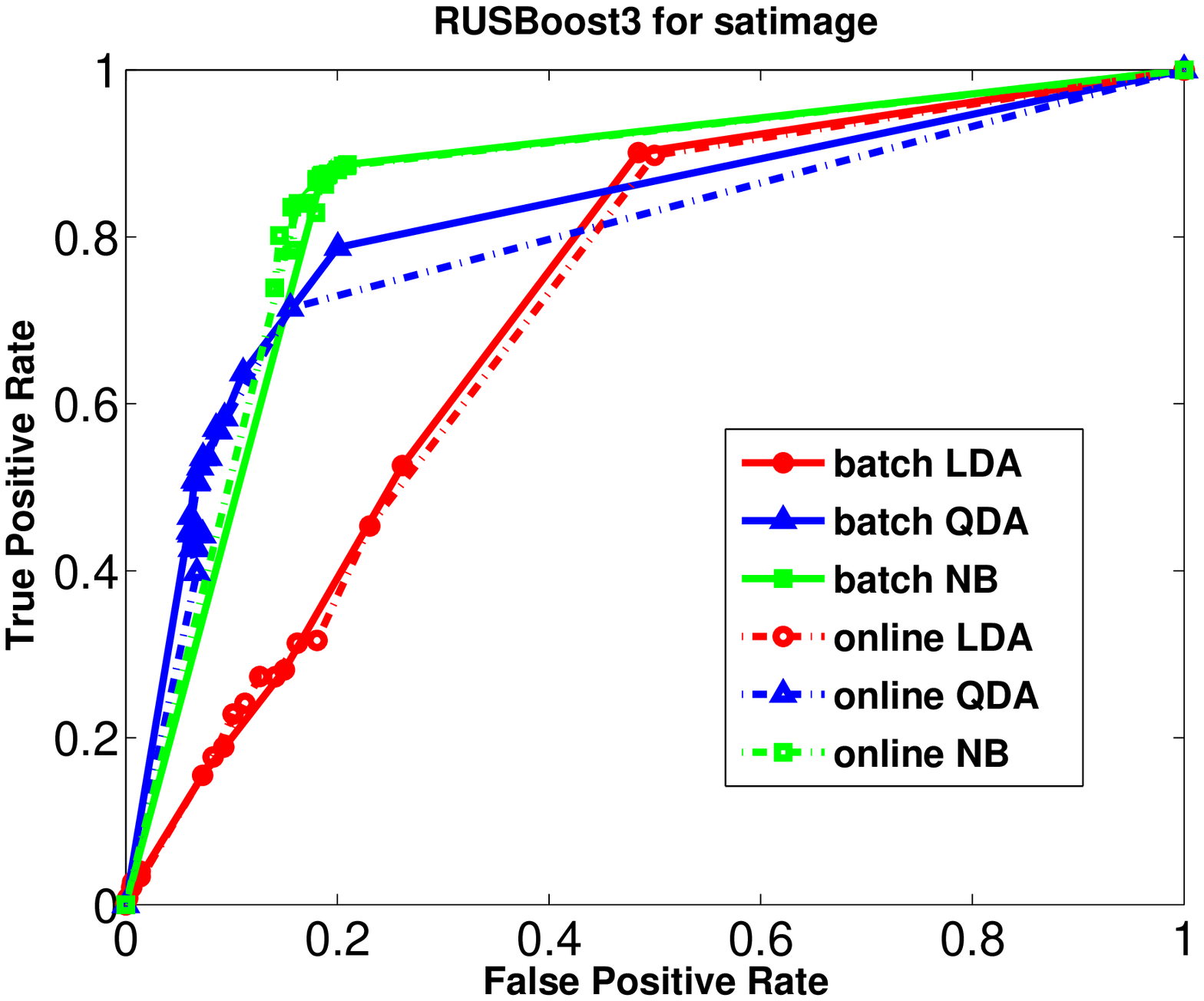}}

\subfigure{\includegraphics[width=2in]{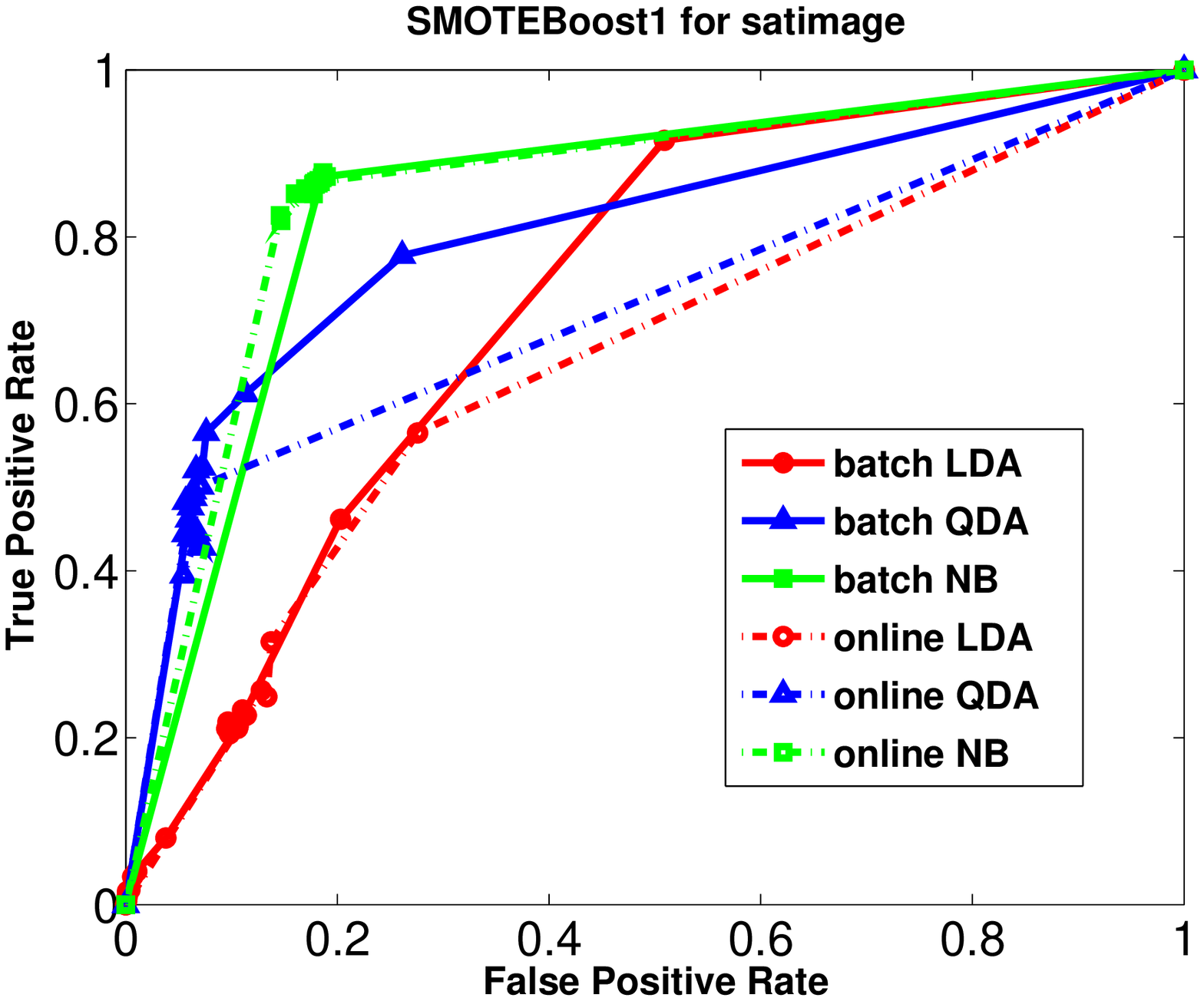}}
\hfil
\subfigure{\includegraphics[width=2in]{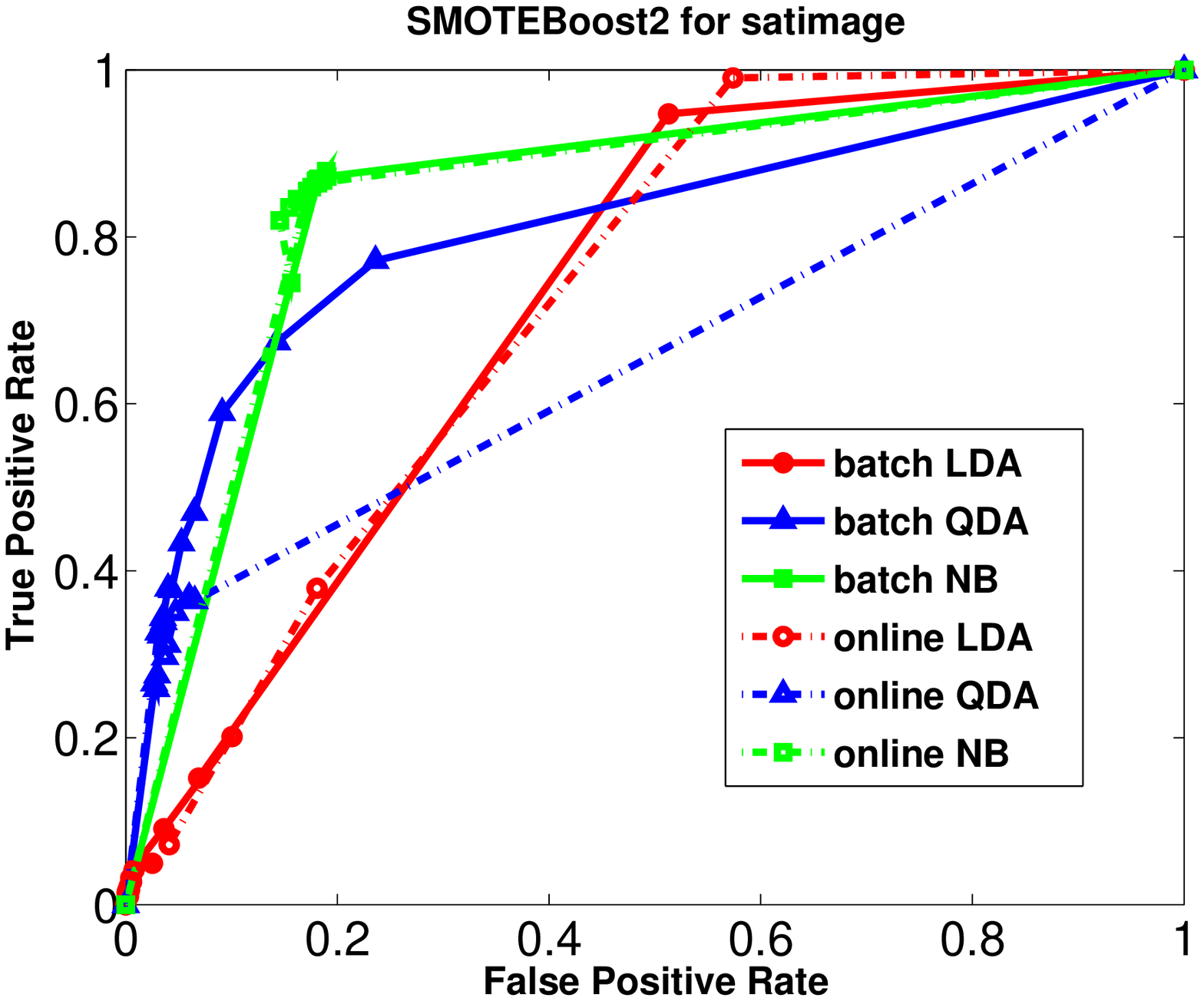}}
\hfil
\subfigure{\includegraphics[width=2in]{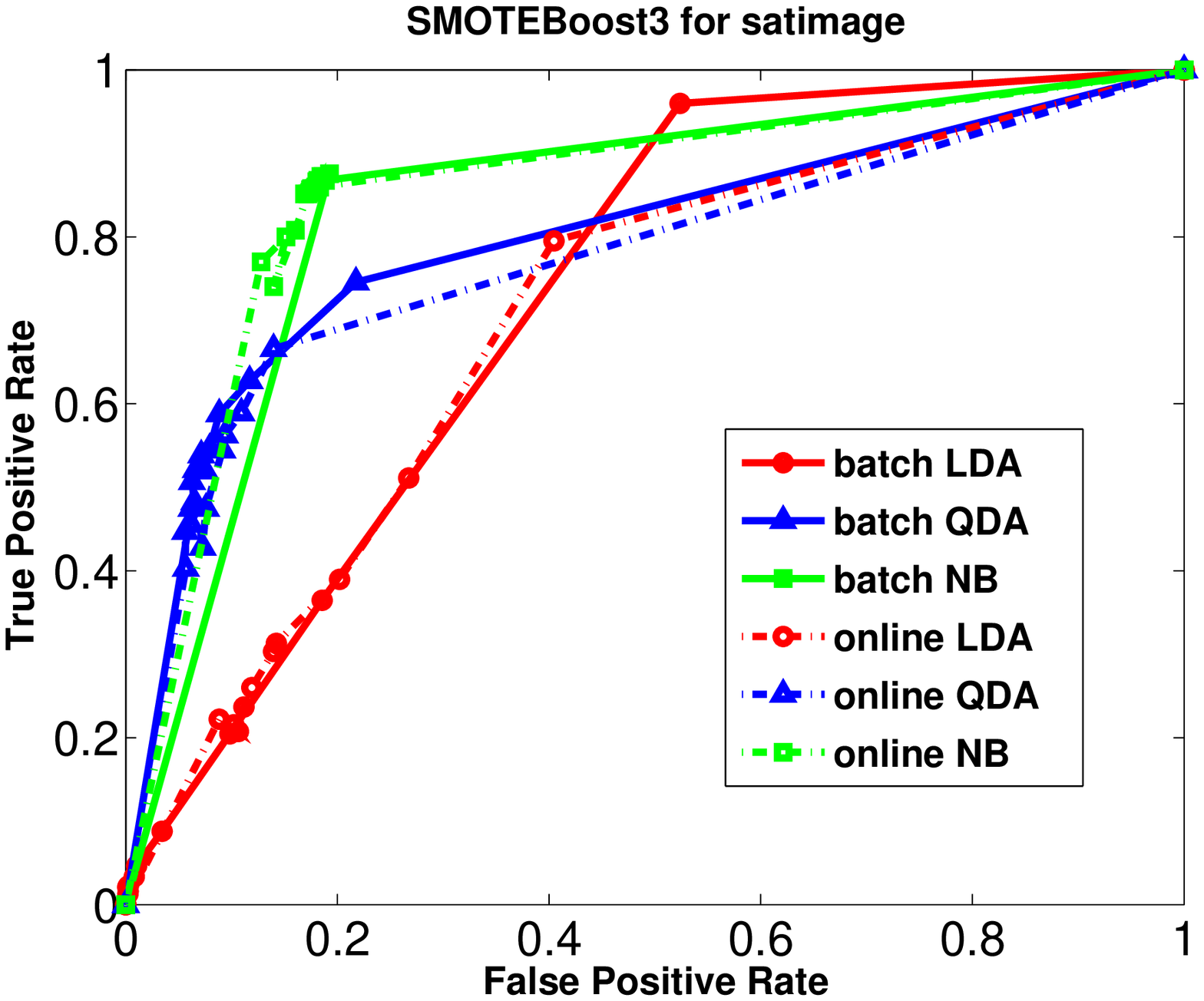}}
\caption{ROCs of different algorithms with different base learner for satimage data set}
\label{fig:satROC}
\end{figure*}

To further investigate the performance and consistency of the algorithms, the box-plot of the absolute difference of AUC between batch and online ensemble algorithms is shown in Fig.~\ref{fig:absDiff}, and the ROCs of different algorithms with different base learners for a particular data set (satimage) are shown in Fig.~\ref{fig:satROC}. It can be observed that the performances of oUOB and oSB is much closer to their batch counterparts than any other algorithm. The ROCs of batch and online bagging algorithms are almost overlapped. Furthermore, this consistency exists across all data sets and all base learners. oAC2 and oCSB2 also show good consistency compared to other boosting algorithms (Note that the 75th percentiles of AdaC2 and CSB2 are smaller than for other boosting-based algorithms). We also observe that oRUS1 demonstrates worst consistency among the algorithms. Therefore, we can conclude that the relative performance of of online cost-sensitive ensemble algorithms are not necessarily consistent with the ones of their batch counterparts. While the performances of batch algorithms are similar to each other, the performances of the online ensembles appears to be mainly determined by the consistency with their batch mode counterparts.

From Fig.~\ref{fig:meanDiff} and Fig.~\ref{fig:absDiff}, it can be observed that LDA and QDA achieve slightly better performance in terms of average AUC than NB yet NB has a slightly better consistency. In addition, oUOB and oSB perform consistently better (in terms of convergence), yet oRUS1 consistently performs worse  than other algorithms across all base learners. Therefore, there is no strong evidence indicating the choice of base learners can significantly affect the performances or consistency of the online ensemble algorithms.

\begin{figure*}
\centering
\subfigure{\includegraphics[width=2.2in]{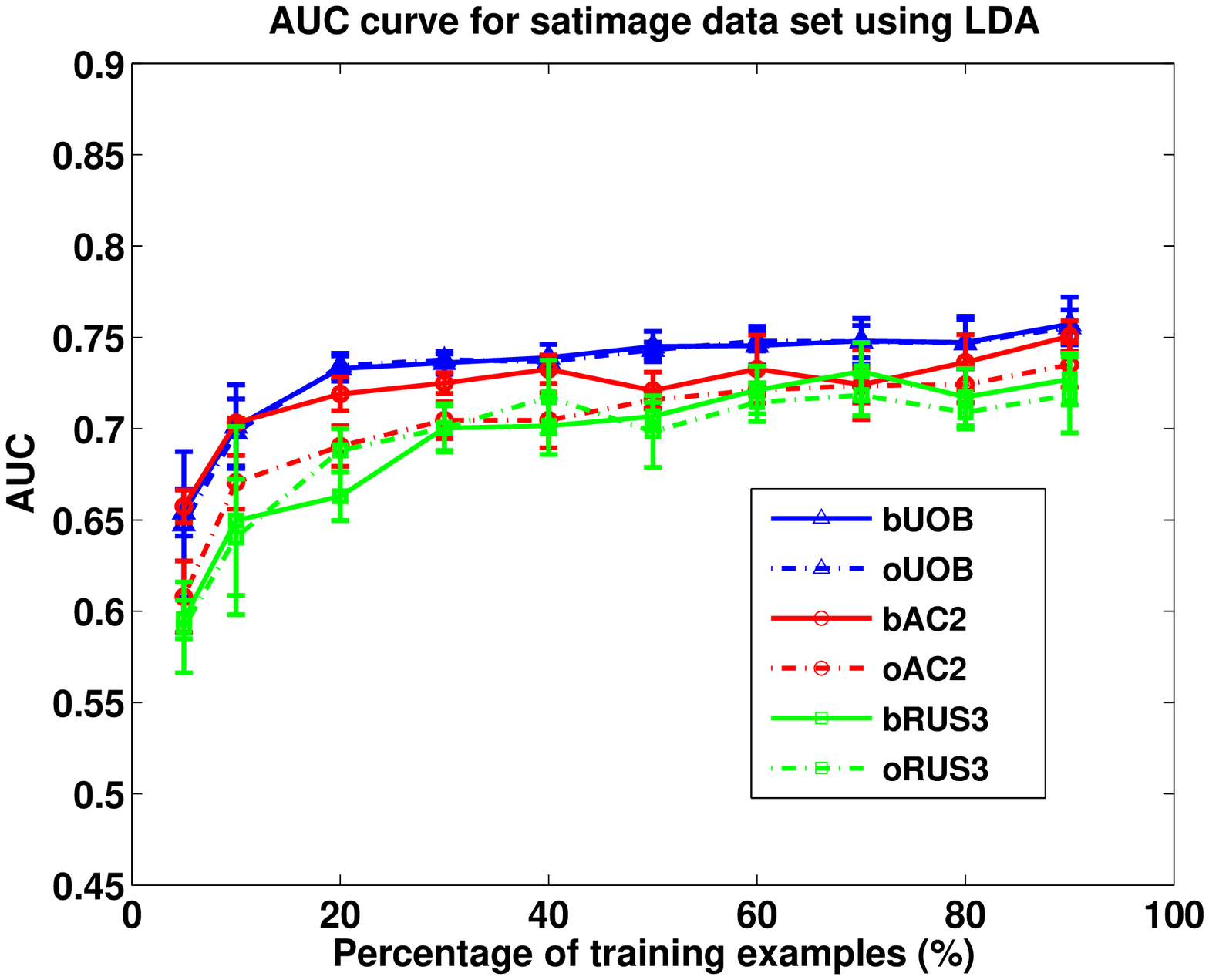}}
\subfigure{\includegraphics[width=2.2in]{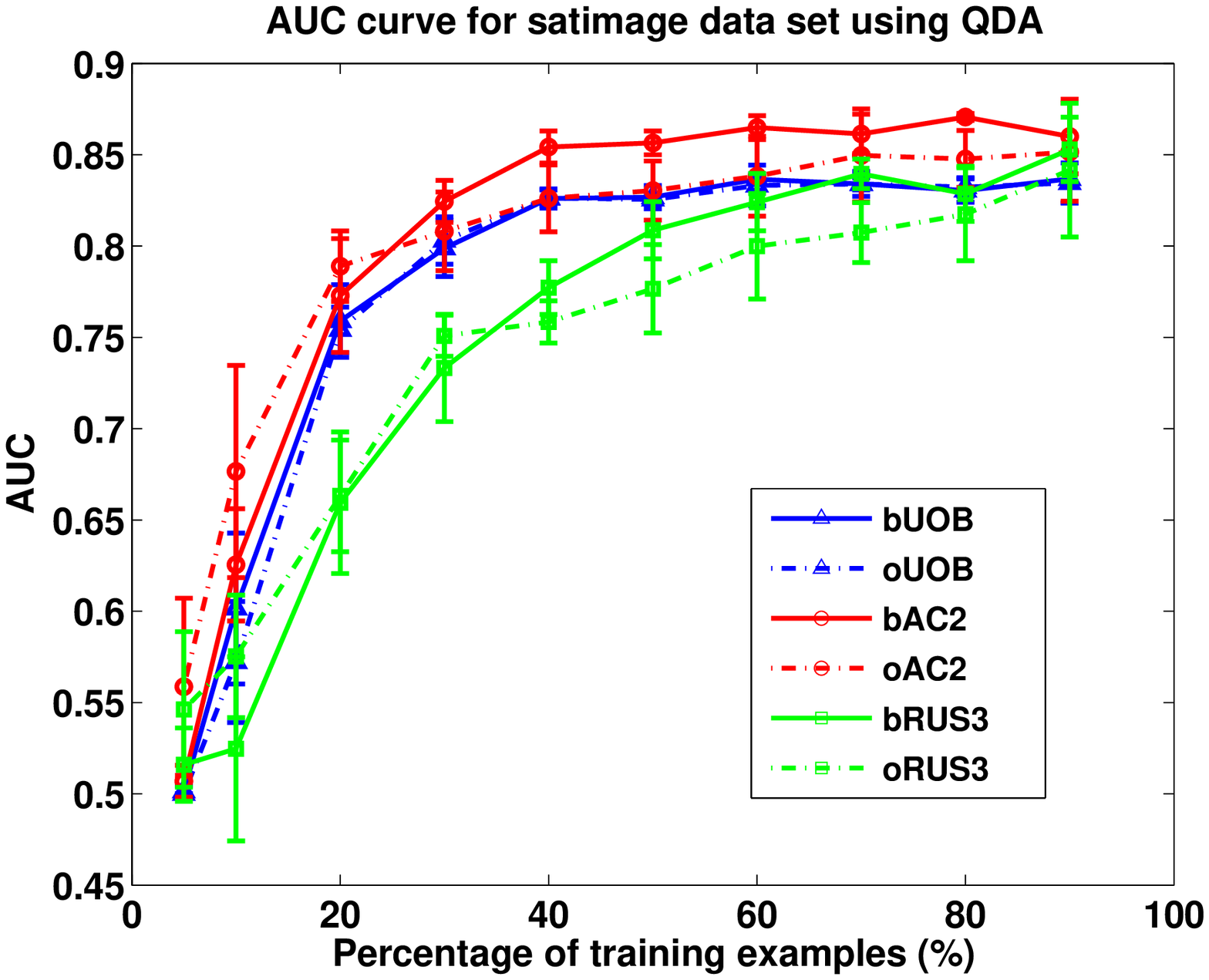}}
\subfigure{\includegraphics[width=2.2in]{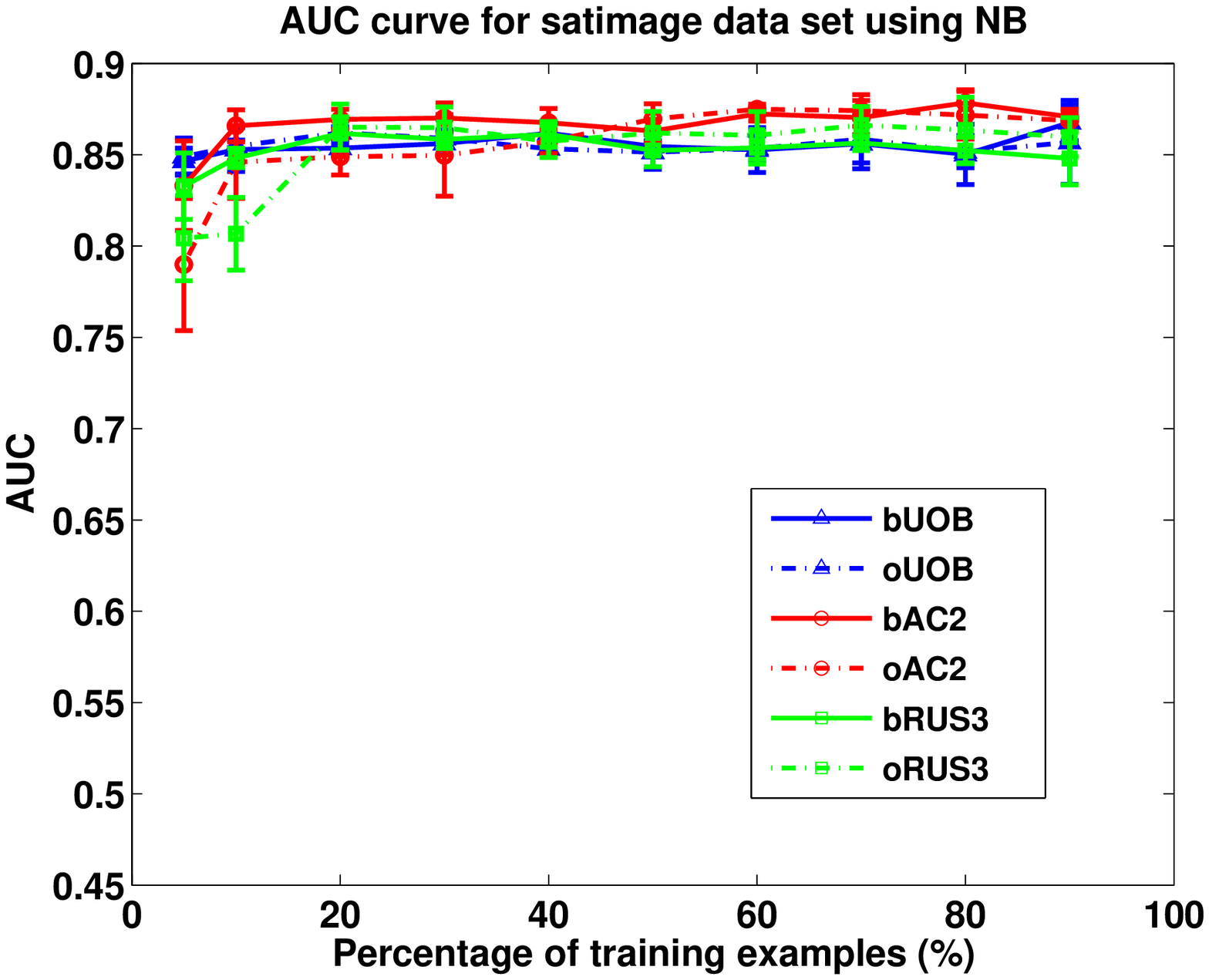}}
\caption{Learning curves with different base learning for satimage data set}
\label{fig:AUCTrend}
\end{figure*}

The last question is the speed of convergence of the proposed algorithms. Because of the better performances of oUOB, oSB, oAC2, oCSB2, and RUS3 in terms of both consistency and AUC, as well as the similarity between oUOB and oSB, oAC2 and oCSB2, we only discuss oUOB, oAC2 and RUS3 as the representatives of online bagging and boosting algorithms. We vary the number of examples of satimage data set from 5\% to 90\% used as training set, and the rest are used as testing set. The experiment for each algorithm is repeated five times, and average AUCs with standard deviation are shown in Fig.~\ref{fig:AUCTrend}. It can be observed that the performance of oUOB converges to bUOB even with a very small amount of training examples. The consistency of online boosting based algorithms cannot be compared with that of UOB, but their performance gets closer to that of their batch counterparts as the number of training example grows, especially when the percentage of training examples is larger than 70\%.

\section{Discussion of the Results}

\begin{figure*}
\centering
\subfigure{\includegraphics[width=3in]{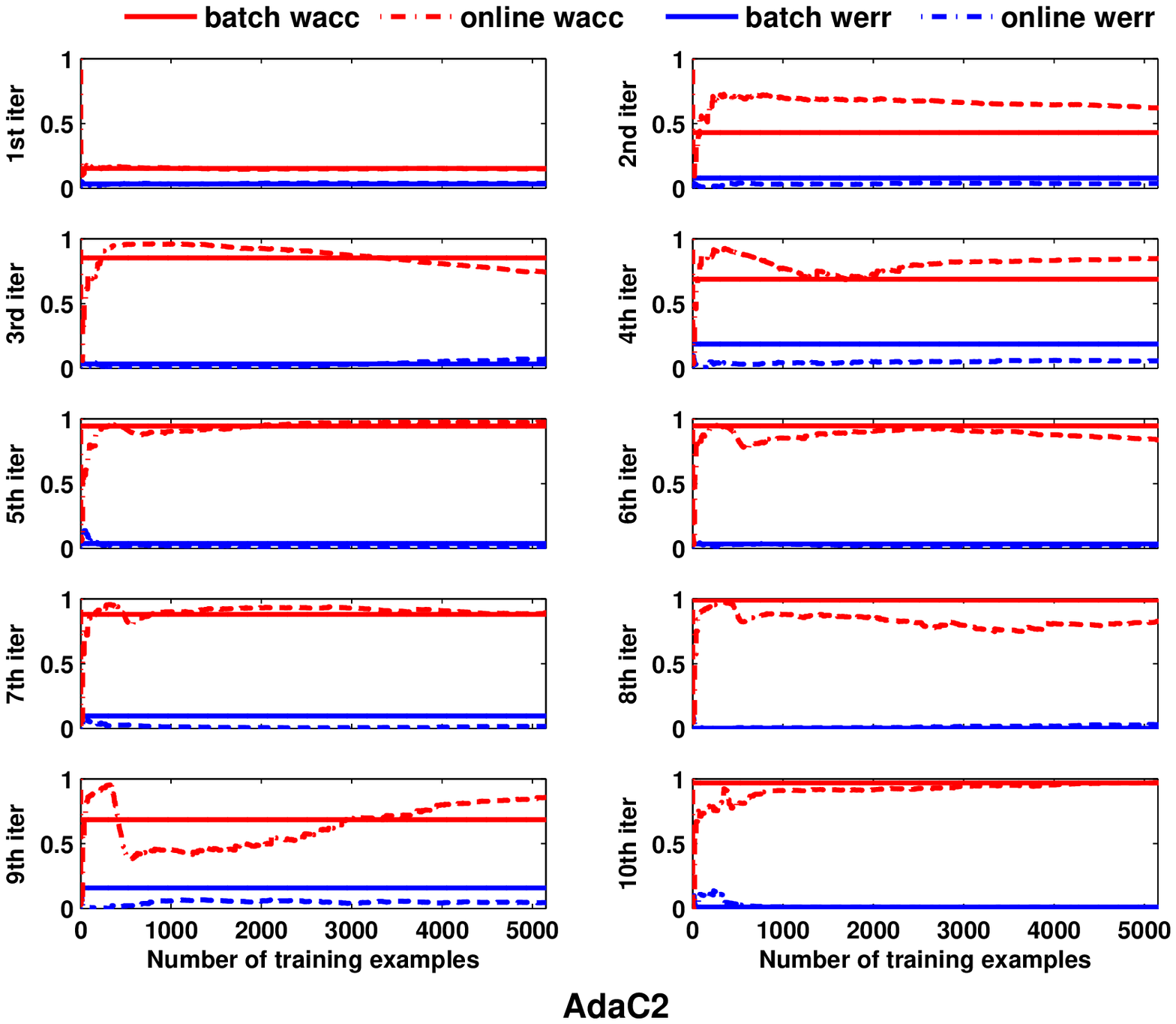}}
\hfil
\subfigure{\includegraphics[width=3in]{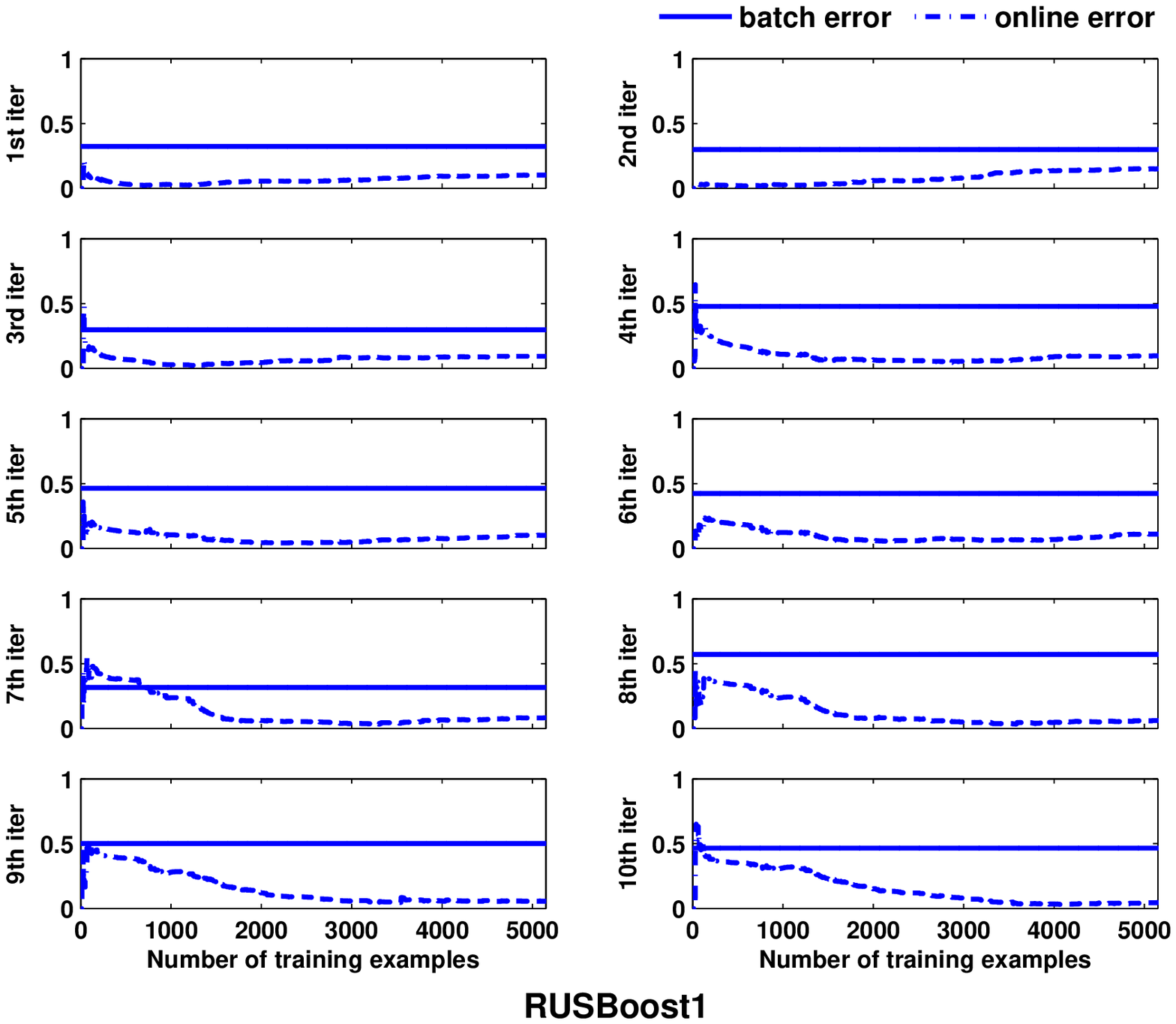}}
\caption{Anytime performances on training examples for satimage data set}
\label{fig:ErrorCurveSatimage}
\end{figure*}

The good consistency of bagging algorithms is not difficult to explain. The approximation of oUOB only comes from approximating a binomial distribution by a Poisson distribution. For AdaC2 and CSB2, there is another approximation, that is, the (weighted) accuracy and error. In the early stages of the learning process, the estimates of these variables can be unreliable, which can deteriorate the performances of the algorithms by improper weight update and base learner combination. For oRUS1, oRUS2, oSBO1, and oSBO2, the performance is further deteriorated by approximating $N^+$ and $N^-$ by $n^+$ and $n^-$. Fig.~\ref{fig:ErrorCurveSatimage} demonstrates the $werr$, $wacc$ of AdaC2 and error of RUSBoost1 on training examples during the learning process with QDA base learners for the satimage data set. The costs are set to be $C_N=0.1$ for AdaC2, and $C=1$ for RUSBoost1. Therefore, the performances in Fig.~\ref{fig:ErrorCurveSatimage} correspond to the left upper points of the ROCs in Fig.~\ref{fig:satROC}. It can be observed that the performance of  AdaC2 converges faster and better than that of RUSBoost1, which can explain their higher consistency for this data set.

\begin{figure*}
\centering
\subfigure{\includegraphics[width=3in]{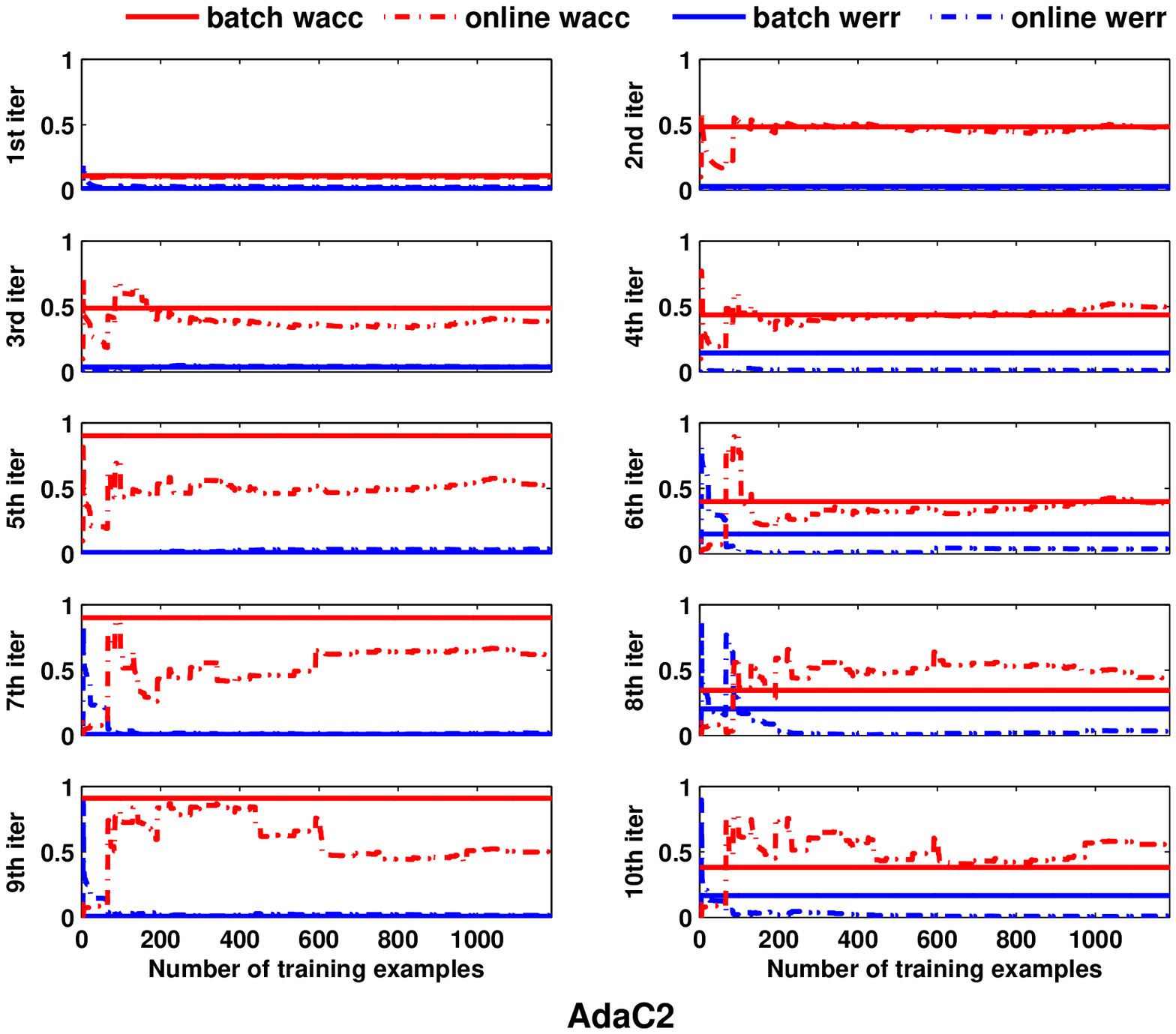}}
\hfil
\subfigure{\includegraphics[width=3in]{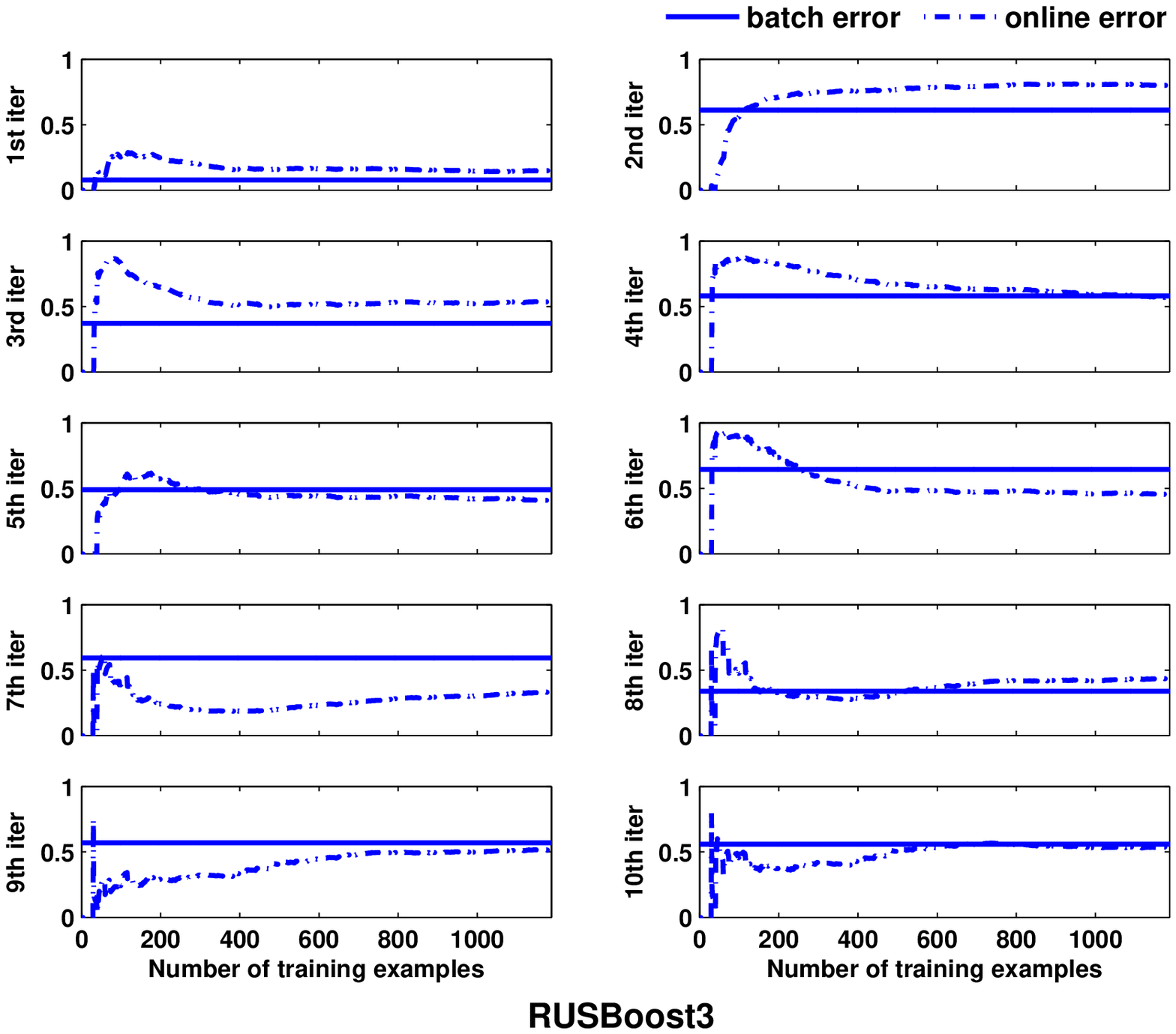}}
\caption{Anytime performances on training examples for yeast6 data set}
\label{fig:ErrorCurveYeast6}
\end{figure*}

In addition, the essential requirement of boosting is that the base learner is at least better than random guess. However, in RUSBoost and SMOTEBoost algorithms, the base learners are trained on the modified data set, but the weights are updated in the same way as the standard boosting algorithm. As a result, base learner trained on modified distribution may have an error larger than 0.5 on the original distribution. On the other hand, AdaC2 and CSB2 do not have this problem since the weights are updated in a way consistent with the target function of the base learner. Therefore, AdaC2 and CSB2 have lower chance to violate the requirement of the algorithms. Usually, for standard boosting, if the $m$th base learner violates the requirements, the algorithm stops and returns the ensemble of $m-1$ base learners. However, in online learning scenarios, the algorithms should not stop since as more training examples pass through the base learners, the error on training examples may become smaller than 0.5. Therefore, to make the experimental setting of batch algorithms consistent with that of online algorithms, the batch boosting algorithms were stopped even when the requirements were violated. As a result, if the error is larger than 0.5, the weights of correctly classified examples increase, whereas the weights of misclassified examples decrease\footnote{For AdaC2 and CSB2, the requirements are $wacc_m>werr_m$ and $\frac{\epsilon_m^2}{1-\epsilon_m}<werr_m$ respectively.}. This improper weight update, propagated through the base learners, may deteriorate the performances of both batch and online algorithms as well as the consistency. Fig.~\ref{fig:ErrorCurveYeast6} shows the anytime performances of AdaC2 and RUSBoost3 on yeast6 with the same experimental setting as in Fig.~\ref{fig:ErrorCurveSatimage}. It is obvious that RUSBoost3 achieves the higher consistency than AdaC2. However, in AdaC2, there is no base learner violating the requirement $wacc>werr$, except for some base learners from the early stage of online learning, while several base learners in RUSBoost3 violate $\epsilon_m<0.5$ in either batch or online (or even both) mode, which may hurt the performance of the algorithm.

In summary, the performances of online cost-sensitive ensemble learning algorithms are largely determined by the consistency to their batch mode counterparts and the performances of the batch ensemble algorithms. Overall, the bagging based algorithms beats all boosting based algorithms in terms of both performance and consistency. AdaC2 and CSB2 achieve comparable performance. The RUSBoost and SMOTEBoost algorithms have worse performance because of their less reliable approximation of the weight update formula, and higher chance of violating the requirements of boosting than AdaC2 and CSB2.

\section{Learning in Non-stationary Environments}

In very recent years, learning of concept drifts from non-stationary imbalanced data streams has attracted increasing attention \cite{gao2007general}, \cite{elwell2011incremental}, \cite{chen2011towards}, \cite{hoens2012learning2}, \cite{hoens2012learning}. Besides the difficulties of standard online cost-sensitive learning, the main challenge of learning in non-stationary environments is that the temporal interval between two successive minority class examples can be quite large, and therefore they can be drawn from different distributions. As a result, it is more difficult to track the concept drifts and the evolution of decision boundary.

As a flexible framework, the online cost-sensitive ensemble learning algorithms can be easily generalized to deal with non-stationarity by leveraging existing algorithms for concept drift. In particular, these methods are developed at the data level, ensemble level, and base learner level. At the data level, the approaches select or reuse examples based on generative models or $k$-nearest neighbors \cite{gao2007general}, \cite{chen2011towards}, \cite{hoens2012learning2} to select instances that are consistent with the current concept. At the ensemble level, the algorithms discard poorly performing base learners and build new ones to adapt to changes in the environments \cite{saffari2009line}, \cite{pocock2010online}. Finally, at the base learner level, the online base learners themselves can be adaptive \cite{anagnostopoulos2012online}. Note that the ensemble level and base learner level approaches are not originally developed for class imbalance problem, but they can be accommodated in our online cost-sensitive ensemble learning framework in a very natural way. In addition, the proposed framework can also combine with drift detection algorithms \cite{baena2006early}.

As an example, here we give a simple yet feasible solution based on online AdaC2 to learning in non-stationary environments. It should be emphasized that the purpose here is to demonstrate the flexibility of the proposed framework for dealing with non-stationarity and suggest the possible research directions in future work, rather than develop sophisticated algorithms. Therefore, the proposed method is quite straightforward and intuitive. The main idea is to put more weights on recent examples than past ones by using \emph{forgetting factors}. For example, the mean $\mu$ and $\Sigma$ of a Gaussian distribution can be incrementally estimated as follows \cite{anagnostopoulos2012online}:
\begin{align*}
  & \mu_n = \left(1-\frac{1}{t_n}\right)\mu_{n-1}+\frac{1}{t_n}x_n, \quad x_0 = 0 \\
  & \Pi_n = \left(1-\frac{1}{t_n}\right)\Pi_{n-1}+\frac{1}{t_n}x_nx_n^T, \quad \Pi_0 = 0 \\
  & \Sigma_n = \Pi_n - \mu_n\mu_n^T,
\end{align*}
with $t_n = \beta t_{n-1}+1$, where $\beta$ is the forgetting factor. Correspondingly, we can also update the weighted Poisson parameters by $\lambda^{TP} \leftarrow \beta\lambda^{TP} + C_P\lambda$, $\lambda^{TN} \leftarrow \beta\lambda^{TN}+C_N\lambda$, $\lambda^{FP} \leftarrow \beta\lambda^{FP} + C_P\lambda$, and $\lambda^{FN} \leftarrow \beta\lambda^{FN} + C_N\lambda$ in steps 10 -- 16 of Alg.~\ref{alg:onlineAdaC2}

Three artificial data sets are used to evaluate the performance of the proposed method on three types of non-stationarity. The first two are SINE1 and SINE1G which are described in \cite{baena2006early}. SINE1 is an unfrequent abrupt concept drift, noise-free data stream of length $N=4000$. The data set has two relevant features with values uniformly distributed in [0,1]. In the first half of the data stream, points lying below the curve $y=sin(x)$ are classified as positive, otherwise are negative. After the midpoint, the classification is reversed. SINE1G models gradual concept drift in a noise-free data stream. The probability of selecting an example from old concept varies from 1 to 0 over the transition period of length $N=2000$. The third data stream SINE1M is a mixture of the previous two of length $N=4000$. The examples gradually drift from the old concept to the new one during the first half of the data stream. After the midpoint, the concept returns to the old one the process of the first half is repeated once more. In all of these three data sets, the class ratio is 90, which is highly imbalanced.

For each data stream, the experiment is repeated ten times. The average online AUC results ($\pm$ for standard deviation) are shown in Table~\ref{tab:NSresults}, where the "ns-" algorithms are the non-stationary versions obtained by setting $\beta=0.9$  The online AUC is generated by predicting each example before it is used for training the classifier. It can be observed that for all cases and all base learners, the simple modification of online AdaC2 can improve the overall performance, especially for SINE1. In summary, the preliminary results are encouraging, indicating that our proposed cost-sensitive online ensemble framework is sufficiently flexible to combine with other existing methods to deal with non-stationarity.

\begin{table}[h]
\setlength{\belowcaptionskip}{10pt}
\caption{Average Online AUC}
\label{tab:NSresults}
\centering
\begin{tabular}{llll}
\hline
            & SINE1            & SINE1G           & SINE1M \\
\hline
LDA         & .7747$\pm$.0408  & .5716$\pm$.0975  & .5547$\pm$.0435\\
nsLDA       & .8885$\pm$.0157  & .5833$\pm$.0680  & .6348$\pm$.0372 \\
QDA         & .7434$\pm$.0444  & .5439$\pm$.0607  & .5214$\pm$.0339 \\
nsQDA       & .8720$\pm$.0229  & .5944$\pm$.0637  & .6048$\pm$.0355 \\
NB          & .5275$\pm$.0309  & .4904$\pm$.0356  & .5174$\pm$.0344 \\
nsNB        & .7270$\pm$.0436  & .5498$\pm$.0598  & .5571$\pm$.0368 \\

\hline
\end{tabular}
\end{table}

\section{Conclusions}

A novel online cost-sensitive ensemble learning framework is proposed in this paper, which generalizes a number of batch cost-sensitive ensemble learning algorithms to their online version. Based on this framework, online learning and cost-sensitive learning are bridged together. The proposed framework is used to derive the online extensions of UnderOverBagging, SMOTEBagging, AdaC2, CSB2, RUSBoost, and SMOTEBoost. We demonstrate theoretically and experimentally that the proposed online cost-sensitive algorithms perform comparably with their batch mode counterparts. In particular, the bagging based algorithms achieve better performance than boosting based algorithms in terms of both consistency and AUC value. AdaC2 and CSB2 also achieve comparable performance. In addition, the proposed framework can also be combined with algorithms for concept drift to learn in non-stationary environments in a very natural way. The encouraging preliminary results on artificial data sets demonstrate that with simple extensions, the proposed algorithms can accommodate a variety of drift scenarios, regardless of whether it is abrupt, gradual, or even a mixture of two. Future work includes designing more efficient and effective algorithms for non-stationary environments with class imbalance and evaluating their performances more extensively, and applying the proposed algorithms to realistic large-scale problems.

\section*{Acknowledgment}

The authors gratefully acknowledge financial support from the Natural Sciences and Engineering Research Council of Canada (NSERC), Discovery grants program, and the Canadian Institutes for Health Research (CIHR).

\bibliographystyle{IEEEtranS}
\bibliography{ref}

\end{document}